\documentclass{article} % For LaTeX2e
\usepackage{colm2024_conference}
\colmfinalcopy
\usepackage{hyperref}

\usepackage{microtype}
\usepackage{graphicx}
\usepackage{subfigure}
\usepackage{booktabs} % for professional tables
\usepackage{hyperref}
\usepackage{subcaption}
\hypersetup{
    colorlinks=true,
    linkcolor=blue,
    filecolor=magenta,      
    urlcolor=cyan,
    pdftitle={Overleaf Example},
    pdfpagemode=FullScreen,
    }
\usepackage{url}
\usepackage{graphicx} % Required for inserting images
\usepackage{algorithm}
\usepackage{algorithmic}
\usepackage{multirow}

\usepackage{xspace}

\usepackage{wrapfig}
\usepackage{pifont}
\usepackage{comment}
\usepackage[inline]{enumitem}

\usepackage{grffile}
\usepackage{xcolor}
\usepackage{color}

\usepackage{newtxmath}
\newcommand{\DL}[1]{\iftoggle{comment}{\textcolor{magenta}{[DL: #1]}}{}}

\newcommand{\AX}[1]{\iftoggle{comment}{\textcolor{magenta}{[AX: #1]}}{}}

\newcommand{\hao}[1]{\iftoggle{comment}{\textcolor{orange}{[Hao: #1]}}{}}

\definecolor{forestgreen}{rgb}{0.13, 0.55, 0.13}

\newcommand{\diag}{\mathrm{diag}}

\newcommand{\vQ}{\mathbf{Q}}
\newcommand{\vK}{\mathbf{K}}
\newcommand{\vV}{\mathbf{V}}
\newcommand{\vdQ}{\mathbf{dQ}}
\newcommand{\vdK}{\mathbf{dK}}
\newcommand{\vdV}{\mathbf{dV}}
\newcommand{\vS}{\mathbf{S}}
\newcommand{\vdS}{\mathbf{dS}}
\newcommand{\vP}{\mathbf{P}}
\newcommand{\vdP}{\mathbf{dP}}

\newcommand{\vO}{\mathbf{O}}
\newcommand{\vdO}{\mathbf{dO}}

\newcommand{\vZ}{\mathbf{Z}}

\newtoggle{comment}
\toggletrue{comment}

\newtoggle{todo}
\toggletrue{todo}

\newcommand{\sysname}{\textsc{DistFlashAttn}\xspace}
\newcommand{\attnname}{\textsc{DistFlashAttn}\xspace}

\usepackage{amsmath}
\usepackage{mathtools}

\usepackage{amsthm}

\usepackage[capitalize,noabbrev]{cleveref}

\theoremstyle{plain}

\theoremstyle{definition}

\theoremstyle{remark}

\usepackage[textsize=tiny]{todonotes}

\title{\attnname: Distributed Memory-efficient Attention for Long-context LLMs Training}

\author{
Dacheng Li\hspace{0.5mm}\thanks{Authors contributed equally.}\hspace{1.5mm}$^b$ \And
Rulin Shao\hspace{0.5mm}$^{*w}$ \hspace{1.5mm} \And
Anze Xie\hspace{0.5mm}$^s$ \And
Eric P. Xing\hspace{0.5mm}$^c$  \AND
Xuezhe Ma\hspace{0.5mm}$^u$  \And
Ion Stoica\hspace{0.5mm}$^b$  \And
Joseph E. Gonzalez\hspace{0.5mm}$^b$  \And
Hao Zhang\hspace{0.5mm}$^s$ \And \ \\
\normalsize$^b$ UC Berkeley \hspace{2.5mm} 
$^w$ University of Washington  \hspace{2.5mm} 
$^s$ UCSD \hspace{2.5mm} 
$^c$ CMU \hspace{2.5mm} 
$^u$ USC
}

\begin{document}

\maketitle

\begin{abstract}
% Increasing the context length of large language models (LLMs) unlocks fundamentally new capabilities, % However, longer context brings up per data memory footprint, where popular data parallelism solutions can not address (e.g. HuggingFace with fully sharded data parallelism). On the other hand, model parallelism systems (e.g, Megatron-LM) resorts to a combination of tensor model parallelism and sequence parallelism to scale to longer context. 
% Extending the context length of large language models (LLMs) introduces new capabilities. %\joey{Make this first sentence a bit more exciting.}
%Na\"{\i}vely training long-context LLMs requires memory that grows quadratically in the sequence dimension.  
%FlashAttention provided a solution to this quadratic memory scaling problem in the single machine setting but 
%the FlashAttention kernel has not yet been extended to the distributed setting.
% memory efficient attention mechanisms have not been extended to the distributed setting. 
%has not yet been extended to the distributed setting.
FlashAttention~\citep{dao2023flashattention} effectively reduces the quadratic peak memory usage to linear in training transformer-based large language models (LLMs) on a single GPU. In this paper, we introduce \attnname, a distributed memory-efficient attention mechanism optimized for long-context LLMs training. We propose three key techniques: token-level workload balancing, overlapping key-value communication, and a rematerialization-aware gradient checkpointing algorithm.
%\attnname is highly optimized for long-context LLMs training through our novel solutions to key chanlleges around token-level work imbalance, key-value communication, and a rematerialization-aware gradient checkpointing strategy.
%Developing a distributed FlashAttention kernel and employing it to support long-context LLMs training introduces several key challenges around token-level work imbalance, key-value communication, and a rematerialization-aware gradient checkpointing strategy.
%\hao{The first sentences are too long for background. Make it shorter and more crispy? Maybe you can just say vanilla flash attention cannot scale.}
%However, training long-context LLMs is severely limited by the available memory on each GPU due to the lack of a distributed implementation of FlashAttention.
% but also significantly expands their memory requirements during training. % , presenting challenges to current methodologies. 
%To address these challenges, we introduce \attnname, a memory-efficient design of sequence parallelism that is highly optimized for long-context causal language modeling.\hao{this sentence first says it is Dist Flash attention, but later it says it is a memory-efficient design of sequence parallelism. The connection between flash attention and sequence parallelism is unclear here.} 
We evaluate \attnname on Llama-7B and variants with sequence lengths from 32K to 512K. \attnname achieves 8$\times$ longer sequences, $4.45 - 5.64\times$ speedup compared to Ring Self-Attention, $2-8\times$ longer sequences, $1.24- 2.01\times$ speedup compared to Megatron-LM with FlashAttention. It achieves $1.67 \times$ and $1.26-1.88\times$ speedup compared to recent Ring Attention and DeepSpeed-Ulysses.
%Anonymous codes are available at \url{https://anonymous.4open.science/r/dist-fa}.
Code is available at \url{https://github.com/RulinShao/LightSeq.}

% \\todo{Regenerate this url to exclude lightseq name}
\end{abstract}

\section{Introduction}
Large language models (LLMs) capable of processing long context have enabled many novel applications, such as generating a complete codebase~\citep{osika2023gpt} and chatting with long documents~\citep{li2023long}. Yet, training these LLMs with long sequences significantly increases activation memory footprints, posing new challenges.

Contemporary approaches to manage the high memory demands of long-context LLMs training involve either reducing activation memory on a single device or partitioning and distributing the sequences across multiple devices. 
Memory-efficient attention~\citep{dao2022flashattention, dao2023flashattention, rabe2021self} represents the former, which reduces the peak memory usage of attention operations on a single device.
% \ion{We are saying that here we discuss about single-machine solutions, so why the next sentence?}
% \DL{We intented to review existing techniques in Paragraph 2,3, and also quickly point out why either of them is not enough by itself. }
%\DL{I propose adding this sentence because people will think "OK we can use Megatron with Flash Attention, why it doesn't work?" I think it is necessary we provide some quick intuition here. : or stitched with other distributed system, an often non-ideal solution as we have leant in past system development~\footnote{In~\S~\ref{sec:exp_baseline} we discuss in depth the solution of stitching FlashAttention with a popular distributed system, Megatron-LM.}~\citep{moritz2018ray}}.
%\todo{(Someone) Review this sentence.}
%since there is no distributed attention mechanism to partition sequence memory costs across machines, memory-efficient attention system has been limited to sequence lengths that fit within a single machine. 
Despite their effectiveness, the absence of a distributed extension limits their application to sequence lengths that a single device can accommodate. 
Naively combining it with existing tensor or pipeline parallelisms~\citep{shoeybi2019megatron}) leads to excessive communication (\S~\ref{sec:comm_mem_analysis}) and cannot scale with sequence length (\S~\ref{sec:exp}).
% For instance, this limitation becomes evident when attempting to distribute attention heads across different devices, e.g., combining with model parallelism (\S\ref{sec:exp_scale}).
On the other hand, sequence parallelism systems, Ring Self-Attention~\citep{li2021sequence} and Ring Attention~\citep{liu2023ring}, distribute the activations of a long sequence across multiple devices,
%partitions a full sequence and assigns subsequences to multiple machines across a distributed cluster. 
but they lack support for memory-efficient attentions (e.g., FlashAttention) or scheduling optimizations, making them inefficient in training long sequences (\S~\ref{sec:exp_rsa}). 
% \rulin{removing the previous explanation since we will discuss the challenges again below.}
% and do not effectively address the workload imbalances inherent in causal attention (i.e., later tokens require more computation than earlier tokens). 
%As we scale the context length and the number of devices, existing sequence parallelism causes high communication time and suboptimal device utilization.
% \hao{I improved your language, but not content. I feel we are still talking the drawbacks of these two categories at a very high level, without clearly articulating the exact issues.}

This paper introduces \attnname to extend the advantages of FlashAttention~\citep{dao2023flashattention} to the distributed setting while maintaining high GPU utilization and low communication overhead. 
\attnname efficiently distributes token chunks across multiple devices, while maintaining the IO-aware benefits of memory-efficient attention.
%leveraging the block-wise computation feature of memory-efficient attention to communicate a minimum number of tensors (i.e., key-value tensors and softmax statistics~\citep{milakov2018online}) required for the current block computation. 
We identify three key challenges in achieving high GPU utilization on distributed FlashAttention design for long-context LLMs and propose three optimizations to addgress them.
%In addition, we identify three challenges that prevent a vanilla design from achieving high GPU utilization and propose three optimizations to address them.

\begin{figure}[t!]
\begin{center}
  \includegraphics[width=\linewidth]{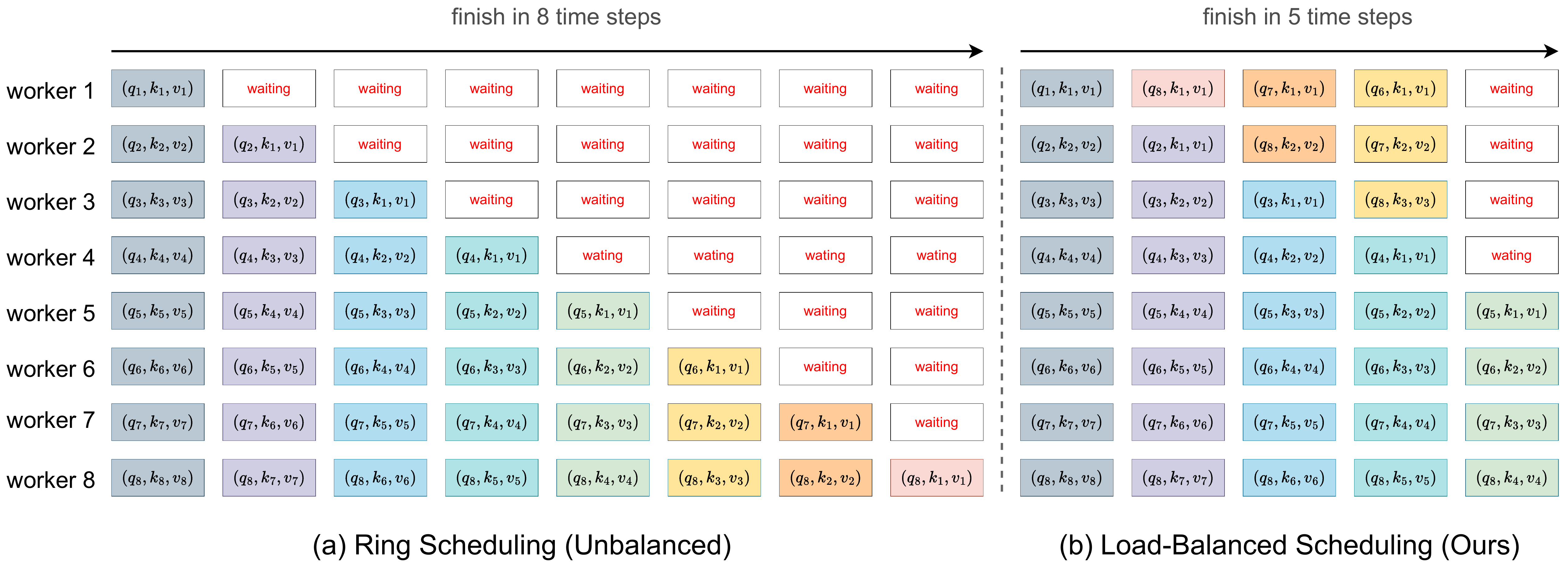}
  \vspace{-0.6cm}
  \caption{
  Per-worker workload at different time steps in (a) ring scheduling~\citep{li2021sequence} and (b) the proposed load-balanced scheduling in an 8-worker scenario. The causal attention introduces a quadratic work dependency on the prefix of each token, where workers assigned earlier tokens remain idle while waiting for workers with later tokens. The idle fraction of the ring scheduling is $\frac{P^2-P}{2P^2}$, asymptotically $\frac{1}{2}$ when scaling to more number of workers. The idle fraction of the proposed load-balanced scheduling is $\frac{1}{2P}$ when $P$ is even and $0$ when $P$ is odd, asymptotically $0$ when scaling to a larger number of workers.
  }
  \label{fig:load_balance}
  \vspace{-4mm}
\end{center}
\end{figure}

The first challenge is the token-level workload imbalance caused by causal language modeling.
As shown in Figure~\ref{fig:load_balance} (a), the causal attention introduces a quadratic work dependency on the prefix of each token. This leads to workers assigned earlier tokens to remain idle while waiting for workers with later tokens to complete, lowering the GPU utilization almost by half. We address this challenge by introducing a load-balancing schedule that routes the extra attention computation of later tokens to those idle workers (\S~\ref{sec:scheduling}). This optimization yields twice throughput of the unbalanced version as shown in Figure~\ref{abl:load_overlap}.

The second challenge is the prohibitive communication overhead.
When tokens are distributed to multiple machines, these machines need to communicate key-value tensors and softmax statistics to jointly compute the global attention. The communication volume is nontrivial, leading to large communication overhead, which grows with the context length. By leveraging the attention dependencies, we propose a scheduling technique that overlaps communication and computation by pre-fetching tensors. This successfully hides communication overhead inside the computation time, resulting in a $1.32\times$ end-to-end speedup (Figure~\ref{abl:load_overlap}) compared to a non-overlapping version.

The third challenge is the high computation overhead due to the re-computation in gradient checkpointing~\citep{chen2016training}. 
Gradient checkpointing effectively trades computation for memory by selectively storing intermediate activations (e.g., the inputs of every layer) and recomputing on-the-fly during the backward pass. It has become a standard technique in the training of long-context LLMs to accommodate the prohibitive activation memory~\citep{zheng2023judging}. However, the recomputation of the FlashAttention causes a high computation overhead in long sequences where the attention dominates the computation time. In \S~\ref{sec:checkpointing}, we show the recomputation of FlashAttention is unnecessary for its backward pass and propose a novel gradient checkpointing strategy to avoid it. Our new strategy results in an 1.31$\times$ speedup (\S~\ref{sec:ablation}) without introducing any numerical difference. 

Our main contributions are:
% \rulin{todo: compat\&shorten}
\begin{enumerate}
    \item %We design \sysname, a long-context LLM training prototype based on sequence-level parallelism. 
    We develop \attnname, a distributed, memory-efficient, exact attention mechanism with sequence parallelism. 
    % \item 
    We propose new optimization techniques to balance the causal computation workloads and overlap computation and computation to increase GPU utilization and reduce communication overhead for training long-context LLMs.
    We also propose a rematerialization-aware gradient checkpointing strategy that eliminates redundant forward recomputation of FlashAttention.
    %in de facto gradient checkpointing without any numerical difference.
%    bypasses the recomputation of FlashAttention forward passes in traditional gradient checkpointing without any numerical difference.
    \item We perform comprehensive evaluation of~\sysname on LLaMA models, against four strong distributed systems. ~\sysname supports $8\times$ longer sequences with $5.64\times$ compared to Ring Self-Attention, $2-8\times$ longer sequences with $1.24-2.01\times$ speedup compared to Megatron-LM. ~\sysname achieves $1.67\times$ and $1.26-1.88\times$ speedup compared to Ring Attention and DeepSpeed-Ulysses. 
    %\item We further demonstrate that \attnname poses no limit on the maximum sequence length, in contrast to tensor and pipeline model parallelism that cannot scale beyond number of heads.
\end{enumerate}

%\joey{Scaling transformers has required significant advances in distributed training techniques.  
%However most existing work has focused on scaling the model size (e.g., depth) and not supporting long context training.
%As a consequence, when training on long-context data, increasing the number of machines does not reduce the memory requirements of attention. 
%While methods like FlashAttention have been instrumental in reducing memory cost on a single machine, there is no distributed attention mechanism to partition sequence memory costs across machines.}

\section{Related work}
\vspace{-1mm}
\paragraph{Memory-efficient attention.} \cite{dao2022flashattention} and \cite{xFormers2022} propose to use an online normalizer~\citep{milakov2018online} to compute the attention in a blockwise and memory-efficient way. It reduces peak memory usage by not materializing large intermediate states, e.g. the attention softmax matrix.
% or the up projection matrix output of the MLP layers\rulin{I'm confused why we talk about MLP in the memory-efficient *attention* paragraph, so I removed it.}~\citep{liu2023blockwise}. 
% Instead, the attentions are computed in smaller blocks and only the final activation are stored. In the backward pass, the intermediate states need to be recomputed.\rulin{removing this because I think we don't need to describe this detail again in the related work section}
% Besides, \joey{besides is an awkward transition in general.}
In addition, research on sparse attention computes only a sparse subset of the attention score, which also reduces the memory footprints yet may lead to inferior performance~\citep{beltagy2020longformer, sun2022length,zaheer2020big}. In this work, we limit our scope to exact attention.
\vspace{-3mm}
\paragraph{Sequence parallelism and ring attention}
Ring Self-Attention~\citep{li2021sequence} is among the first to parallelize Transformers in the sequence dimension. However, its distributed attention design is not optimized for causal language modeling and incompatible with memory-efficient attention, which are crucial for long-context LLM training. Ring Attention~\citep{liu2023ring} proposes to compute distributed attention in a memory-efficient blockwise pattern. However, it is also not optimized for causal language modeling, leading to 2$\times$ extra computation. \sysname optimizes for both memory-efficient attention and causal language modeling. More recently, DeepSpeed Ulysses~\citep{jacobs2023deepspeed} proposes a hybrid parallelism strategy. It computes distributed attention in the tensor model parallelism to address these two problems and utilizes sequence parallelism elsewhere~\citep{shoeybi2019megatron}. We provide head-to-head comparison in Table~\ref{tab:comp_ds}.
\vspace{-3mm}
\paragraph{Model Parallelism and FSDP}
Tensor Model parallelism~\citep{korthikanti2023reducing} partitions model parameters and also distributes the activation in parallel LLM training. Pipeline model parallelism~\citep{huang2019gpipe} also partitions the activations. However, it applies high memory pressure to the first pipeline stage. We show in~\S~\ref{sec:exp_megatron} that this leads to a less effective support for long sequences. Thus, we focus on comparing with tensor model parallelism and only consider pipeline parallelism when the number of heads is insufficient for tensor parallelism.
% In particular, tensor parallelism distributes the model weights within each layer across devices and pipeline parallelism distributes different layers across devices. 
% Megatron-LM~\citep{korthikanti2023reducing} proposes a hybrid usage of tensor parallelism and sequence parallelism to better reduce the activation on a single device and is the main baseline of the paper. 
Fully sharded data-parallelism (FSDP)~\citep{zhao2023pytorch, rajbhandari2020zero} distributes optimizer states, gradients, and model parameters onto different devices and gathers them on-the-fly. Our work focuses on reducing the activation memory that dominates in long-context training. Therefore, FSDP is orthogonal to our work.%, and we use~\sysname in tandem with FSDP to further reduce memory acquired by models in experiments. 

%\paragraph{Pipeline parallelism}
\vspace{-3mm}
\paragraph{Gradient checkpointing.}
Gradient checkpointing~\citep{chen2016training} trades computation for memory by not storing activations for certain layers and recomputing them during the forward pass. Selective checkpointing~\citep{korthikanti2023reducing} suggests recomputing only the attention module, as it requires significant memory but relatively few FLOPs (in contexts of smaller length). Checkmate~\citep{jain2020checkmate} finds optimal checkpointing positions using integer linear programming. However, none of these designs have considered the effects of memory-efficient attention kernels, which perform recomputation within the computational kernel to avoid materializing large tensors. In this paper, we demonstrate that by simply altering the checkpointing positions, we can avoid the recomputation of these kernels without introducing any numerical difference.

% As a result, many previous recomputation policies become less effective. 
% In this work, we focus on checkpointing at the boundary of every transformer layer, which is a popular strategy adopted by many current open-sourced projects to accommodate the prohibitive activation memory.

% \pw{nit: use citep and cite consistently in this section}

%\paragraph{Long-context Transformer.}\rulin{add Code-Llama, etc that has heads not divisible by 8, 16; add models like CodeGen and gpt-2 that have less heads; Discuss MQA, GQA.}
%There are many l

% \textbf{Sequence parallelism}
% Each Transformer layer can be viewed as a token mixer and other~\citep{yu2022metaformer}. Essentially, only the attention module poses dependency for different tokens, and all other operations are computed independently in each token. This natural limited dependency gives the idea of paralleling the sequence dimension. 
\begin{figure}[t!]
\begin{center}
  \includegraphics[width=0.8\textwidth]{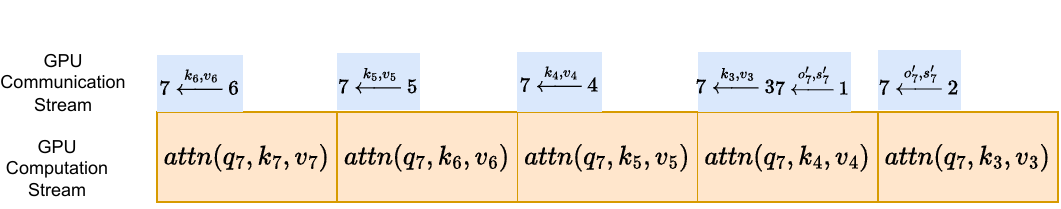}
  \vspace{-0.3cm}
  \caption{Overlap example in the forward pass of worker 7 out of an 8 worker scnerio. For simplicity, "worker p" is denoted as p.}
  \label{fig:overlap}
  \vspace{-6mm}
\end{center}
\end{figure}

\section{Method}
In this section, we first present a distributed memory-efficient attention mechanism that distributes the computation along the sequence dimension, \attnname (\S~\ref{sec:dist_attn}) in its vanilla form. We then introduce two novel optimizations in~\attnname: a load-balanced scheduling strategy for causal language modeling to reduce the computation bubble and an asynchronous communication design that overlaps the communication into computation (\S~\ref{sec:scheduling}). Finally, we propose a new rematerialization-aware checkpointing strategy (\S~\ref{sec:checkpointing}) which effectively cuts off the recomputation time in gradient checkpointing when using~\attnname in long-context training.

\subsection{\attnname: distributed memory-efficient attention via sequence parallelism}
\label{sec:dist_attn}
% \joey{I rewrote the setup here to give more high-level context up-front.}
%The core idea in %/\attnname is % to split a $N$ tokens long input sequence evenly into $P$ subsequences and distributes the computation and a
The goal of \attnname is twofold: (1) distribute a single sequence into multiple workers so they jointly utilize the memory to support a long sequence training; (2) maintain the IO-aware benefits of memory-efficient attention so that training is fast and incurs less memory footprint. In particular, we choose FlashAttention~\citep{dao2023flashattention} as the paradigm. %For ease of illustration, we only show how the forward pass is computed, where the backward pass is designed in similar way.

\textbf{To distribute the long sequence.} \attnname splits the input sequence consisting of $N$ tokens evenly across $P$ workers (e.g. GPUs) along the sequence dimension. Each worker computes and stores the activations of only a subsequence of $N / P$ tokens. Therefore, it supports training $P\times$ longer with $P$ workers than a single-worker FlashAttention.

%the forward and backward pass for only $N / P$ of the $N$ tokens, 

%For modules like the Feed Forward Layer (FFN), Layer Norm (LN), and the embedding layer the tokens can be computed independently without coordination (embarrasingly parallel) and the work is balanced across workers.
% can be embarrassingly parallelized along the sequence dimension 
%Unfortunately, in the attention modules where local tokens may need to attend to remote tokens, coordination is required.
%Unfortunately, for the attention modules where local tokens may need to attend to remote tokens, coordination is required. % and work maybe imbalanced, e.g. in causal language modeling.
% Therefore, we focus on the sequence parallelism design of the attention module where the local tokens may need to attend to the tokens on the remote workers. 
%To address this, each worker collects all the keys and values associated with other tokens and then locally computes the attention following~\citet{dao2023flashattention}.
%To address the memory pressure introduced by collecting all other keys and values, this process is done online by streaming the key and values from workers with earlier tokens to workers with later tokens.
Formally, let $\mathbf{q}_p$, $\mathbf{k}_p$, $\mathbf{v}_p \in \mathbf{R}^{\frac{N}{P}\times d}$ be the query, key and value of the subsequence on the $p$-th worker ($p = \{1, \cdots, P \}$), where $d$ is the hidden dimension. Considering the most prevalent causal attention in LLMs, worker p computes the attention output $\textbf{o}_p$ associated with $\textbf{q}_p$:
%the attention output on the $p$-th worker is: %\rulin{todo: include causal mask in this equation, show that the computation for the p-th worker only needs k,v chunks from 1 to p.}%\rulin{replace this description with an attention equation taking o=softmax($q^T \cdot k1...kP) v1 ... vP$?}\DL{Yep! Good point!}
%\rulin{also note that the index should be put outside the bold}
\begin{equation}
    \mathbf{o}_p = \text{Softmax}(\frac{\mathbf{q}_p[\mathbf{k}_1,...,\mathbf{k}_p]^T}{\sqrt{d}})[\mathbf{v}_1,...,\mathbf{v}_p]
\end{equation}
%where $d$ is the hidden dimension~\citep{vaswani2017attention} .
% \AX{we need to mention what id $d_k$ in the above equation}
\textbf{To maintain the IO-awareness}. Na\"{\i}vely, each worker could gather all the keys and values associated with other subsequences and then locally computes $\mathbf{o}_p$ by invoking the existing single machine FlashAttention.
However, this gathering introduces memory pressure by having to store the full list of keys and values locally, a total size of $\mathbf{R}^{2N\times d}$.
%\rulin{can we list the tensors needed to be gathered in this way? so we can compare our design with it later.}. 

Fortunately, the block-wise nature of the single-worker FlashAttention only requires \textbf{one} block of keys and values in each iteration of its algorithm, %where the presence of the full list of keys and values is not required.
%does not require the presence of the full list of keys and values. Its iterative algorithm only requires one block of 
%It iterates over a block of keys and values, computes partial attention results, and rescale before the next iteration.
%To address this memory pressure, we leverage the online softmax algorithm for attention ~\citep{milakov2018online,rabe2021self, dao2022flashattention}, which 
Leveraging this observation, we compute $\mathbf{o}_p$ iteratively: in each iteration when $r\neq p$, worker p fetches only \textit{one} $\mathbf{k}_r, \mathbf{v}_r$ from a remote worker $r$, It then computes partial attention results based on $\mathbf{q}_p$ and $\mathbf{k}_r, \mathbf{v}_r$ and perform proper rescaling by invoking the single-worker FlashAttention kernel. %, and rescale~\citep{milakov2018online} to get the right output.
%iteratively computes partial attention results between $\mathbf{q}_p$, and only \textit{one} $\mathbf{k}_r, \mathbf{v}_r$ from a remote worker $r$. 
To perform proper rescaling between iterations, each worker also needs to maintain a copy of softmax statistics\footnote{These are statistics $l$ and $m$ in FlashAttention words.} $\mathbf{s}_p \in \mathbf{R}^{\frac{2N}{P}}$. %\rulin{mention the shape of these saved tensors?} 
%and performs a rescaling to update partial results from previous iteration. 
Computing in this iterative manner, each worker also stores the key and value of one subsequence of size $\mathbf{R}^\frac{2N\times d}{P}$, a factor of $\frac{1}{P}$ memory of the na\"{\i}vely design. We refer to \citet{dao2022flashattention} for more details of the single-worker FlashAttention. We denote each iteration of the partial attention result and the rescaling as $attn(\mathbf{q}_{p}, \mathbf{k}_{r}, \mathbf{v}_{r}, \mathbf{s}_{p})$, and present the vanilla~\attnname algorithm in Algorithm~\ref{alg:distflashattn_forward_highlevel_unbalanced}.\
% \todo{Rulin comments: a bit hard to follow if the reader isn't super familiar with FA equations. can we add a minimum equation of FA to at least explain what is the softxmax statistics?}
In Appendix~\ref{sec:attn_in_lightseq}, we show how to implement the ${attn}(\cdot)$ kernel from~\cite{dao2023flashattention} in pseudo-code.

\subsection{Load balanced scheduling with communication and computation overlap}\label{sec:scheduling}
\paragraph{Load-balanced scheduling.} %Causal language modeling objective~\citep{brown2020language,touvron2023llama} 
%is one of the most prevalent objectives for LLMs, 
In causal attention, each token only attends to its previous tokens, i.e. the p-th worker computes ${attn}(\mathbf{q}_p, \mathbf{k}_{r}, \mathbf{v}_{r})$ for all $r \leq p$. This introduces a workload imbalance between workers: a worker with a larger $p$ computes more ${attn}(\cdot)$ (Figure~\ref{fig:load_balance} (a)). Using the scheduling described in~\S~\ref{sec:dist_attn}, the idle fraction is $\frac{P^2 - P}{2P^2}$ ($\rightarrow \frac{1}{2}$ when $P \rightarrow \infty$), which means roughly half of the workers are idle. To reduce this idle time, %(a.k.a., the bubble time), 
we let worker $r_1$ that has finished all its ${attn(\cdot)}$ computations (i.e., the ``helper'') perform attention computation for worker $r_2$ with heavier workload, as shown in Figure~\ref{fig:load_balance} (b).
% with ${attn}(\mathbf{q}_r, \mathbf{k}_{q}, \mathbf{v}_{q})$, where worker $r$ has heavy workload, i.e. $r > \frac{P}{2}$.
%fetch a remote $\mathbf{q}_{r}$ instead of $\mathbf{k}_{r}, \mathbf{v}_{r}$, 

Notably, the ``helper'' $r_1$ needs to communicate softmax statistics and the partial attention output to the original worker $r_2$, so that worker $r_2$ can update its local copy of statistics and output correctly (Algorithm~\ref{alg:distflashattn_forward_highlevel_balanced}). This update function is denoted as ${rescale(\cdot)}$ and
updates the partial output and statistics in the same way as how~\cite{dao2023flashattention} updates results from two block execution. This scheduling gives an average idle time fraction:
%has the same logic as the update function of two consecutive block computation in~\cite{dao2023flashattention}.
%We also provide a figure for $P=8$ in "Load-Balanced Scheduling" of Figure~\ref{fig:load_balance}.
%that have finished their computation fetch a remote query instead of key and value, and compute help compute for a remote worker $r$ (Algorithm~\ref{alg:distflashattn_forward_highlevel_balanced} and "Load-Balanced Scheduling" in Figure~\ref{fig:load_balance}). Importantly, in IO-aware ${attn(\cdot)}$ the softmax statistics are associated with the corresponding query. Thus 

%\todo{(Pass 2): use algorithm box and formula}
\begin{figure}
    \centering
    \includegraphics[width=\linewidth]{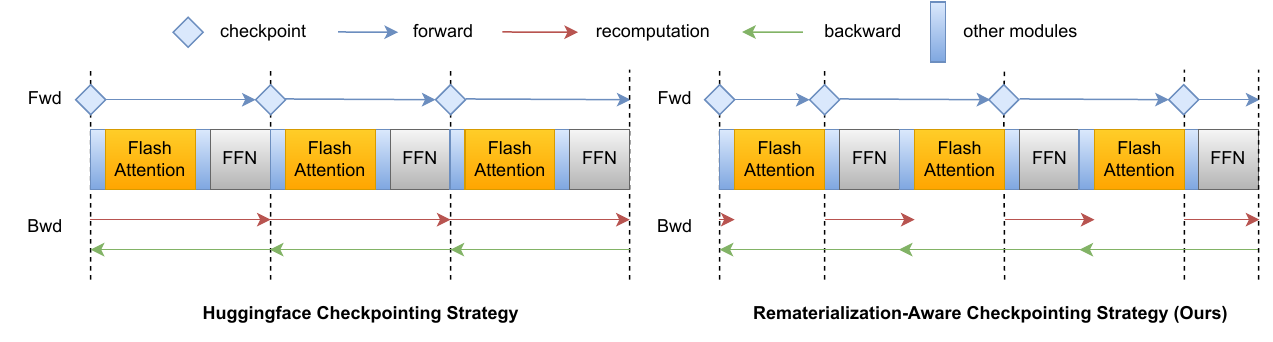}
    \caption{Comparison of HuggingFace gradient checkpointing strategy and our materialization-aware gradient checkpointing strategy. Note that our checkpointing strategy \textbf{saves an entire flash attention forward per layer in recomputation} by simply shifting the checkpoint boundaries without introducing any numerical difference. The checkpointed tensors, i.e., the outputs of FlashAttention, are saved not only for the recomputation of subsequent layers but also the backward computation of the preceding FlashAttention.}
    \label{fig:ckpt}
\end{figure}
\begin{equation}
  X=\left\{
  \begin{array}{@{}ll@{}}
    0, & \text{P is odd} \\
    \frac{1}{2P}, & \text{P is even}
  \end{array}\right.
\end{equation}
Note that when P is even, the idle time is asymptotically $0$ to more workers. We provide an illustration with 8 workers in Figure~\ref{fig:load_balance} and a more detailed one in Appendix~\ref{app:load_balance}. While we focus on the exact attention mechanism, we also discuss sparse patterns in Appendix~\ref{sec:appendix_sparse}.

%as shown in Figure~\ref{fig:load_balance} (``Before Balancing''), in an 8-worker ($P=8$) scenario, the last worker needs to attend to tokens on all other 7 workers, while the first worker is idle after attending to its local tokens, which results in a total idle time of $28$. For instance, we let worker $1$ compute ${attn}(\mathbf{q}_8, \mathbf{k}_1, \mathbf{v}_1)$ and send the result to worker $8$. When the number of workers is odd, the idle fraction is 0. When the number of workers is even, the idle fraction is $\frac{1}{2P}$, which is asymptotically $0$ when scaling to more number of workers.
%We further provide a detailed illustration with time steps of the 8-worker example
%in Appendix~\ref{app:load_balance}.

\paragraph{Communication and computation overlap.}  \attnname relies on peer-to-peer (P2P) communication to fetch $\mathbf{k}_{r}, \mathbf{v}_{r}$ (or $\mathbf{q}_{r}$ in the load-balanced scheduling) from remote workers before computing ${attn(\cdot)}$. However, these communications can be naturally overlapped. To simplify the equations, we use the unbalanced schedule to describe the intuition, while the final~\attnname implementation are equipped with both optimizations. Precisely, these two operations are parallelized:
%\todo{(Pass 2): use formula}
%For instance, When the first worker is computing attention for its local token, it can pre-fetch the next chunk of tokens it needs for the next time step.
\begin{equation}
\begin{aligned}
    \text{Fetch}: \text{worker} \ p
          \xleftarrow{\mathbf{k}_{r+1}, \mathbf{v}_{r+1}} \text{worker} \ r+1 \\
    \text{Compute}: attn(\mathbf{q}_p, \mathbf{k}_{r}, \mathbf{v}_{r}, \mathbf{s}_p)
\end{aligned}
\end{equation}
%worker $p$ can fetch $\mathbf{p}_{r}, \mathbf{v}_{r}$ . 
Thus, in the next iteration, $\mathbf{k}_{r+1}, \mathbf{v}_{r+1}$ are already stored in the memory of worker p, without blocking the next iteration's computation. In modern accelerators, this can be done by placing the attention computation kernel in the main GPU stream, and the P2P communication kernel in another stream, where they can run in parallel~\citep{zhao2023pytorch}. We demonstrate the overlapped scheduling for worker 7 in the 8-worker scenario in Figure.~\ref{fig:overlap}. Empirically, we find this optimization effectively reduces the communication overhead by hiding the communication time inside the computation time (\S~\ref{sec:ablation}).

\subsection{Rematerialization-aware checkpointing strategy}\label{sec:checkpointing}
Gradient checkpointing~\citep{chen2016training} is a de-facto way of training long-context transformers.
% , which trades computation for memory by selectively storing intermediate activations and recomputing others as needed during the backward pass, rather than retaining all activations in memory throughout the process. 
Often, the system uses heuristics to insert gradient checkpoints at the boundary of each Transformer layer~\citep{wolf2019huggingface}. However, with the presence of \citet{dao2022flashattention}, we find the previous gradient checkpointing strategy causes a redundant recomputation of the FlashAttention forward kernel. 
Precisely, when computing the gradient of the MLP layer, \citet{wolf2019huggingface} re-computes the forward of the entire Transformer layer including FlashAttention.
% and save all intermediate activations needed for backward computation. 
During this process, the FlashAttention backward kernel re-computes the softmax block-wisely again to reduce memory usage. 
Essentially, this is because FlashAttention does not materialize the intermediate values during the forward, and recomputes it during the backward, regardless of the re-computation in the outer system level (e.g., the HuggingFace gradient checkpointing~\citep{wolf2019huggingface}). 

To tackle this, we propose to insert checkpoints at the output of the FlashAttention kernel, instead of at the Transformer layer boundary. We use each checkpoint not only for the recomputation of its subsequent modules but also for the backward computation of its preceding FlashAttention module without recomputation. Thus we only need to compute the forward of FlashAttention once, effectively avoiding all recomputations of FlashAttention as shown in Figure~\ref{fig:ckpt}. 

Figure~\ref{fig:attn_time} shows that attention dominates in the forward pass with in long sequences, which indicates our method saves $\sim 0.23\times 32$ (i.e., $\sim 7$) seconds when training a 64K sequence example on Llama-7b on a single machine.
% \rulin{"the local version of flash attention" is a vague definition}. 
In addition, this saves a communication brought by our \attnname forward in the distributed training scenario. We benchmark the end-to-end speedup brought by this materialization-aware checkpointing strategy in \S~\ref{sec:ablation}.

%\subsection{Putting things together}
%\label{sec:put_together}
%\attnname can be easily used as  

%\begin{figure}[h!]
%\begin{center}
%  \includegraphics[scale=0.5]{./figures/attn_time.png}
%  \caption{Time break down of attention versus mlp in a single forward pass, over different sequence lengths.}
%  \label{fig:attn_time}
%\end{center}
%\end{figure}

%\subsection{Communication and Computation Analysis and the Effect of MQA and GQA}
%\rulin{discuss here or not}
%We now present our takeaways:
%\begin{enumerate}
%    \item In long-context, megatron's selective checkpointing is sub-optimal, as (1) the attention FLOP dominates, (2) the MLP memory takes too much, if not checkpointed.
%\end{enumerate}
%\DL{First point overall, sequence length asmptotically dominates the memory, so the saving brought by tensor model parallelism can be ignored.}
\section{Experiments}
\label{sec:exp}
We evaluate~\attnname together with our new checkpointing strategy against alternative distributed approaches for long-context LLMs training. Our primary baseline is Megatron-LM~\citep{shoeybi2019megatron}, used in tandem with FlashAttention, which serves as a robust baseline extensively adopted within the industry. In Appendix~\ref{sec:comm_mem_analysis}, we also show a theoretical analysis on its high communication volume. We also provide a comparison with the previous sequence-parallel system~\citep{li2021sequence}. In addition, we include comparison to recent systems including DeepSpeed-Ulysses and Ring Attention~\citep{jacobs2023deepspeed, liu2023ring}. In the ablation study, we delineate the individual contributions of each component of our methodology, specifically load balancing, computation-communication overlapping, and rematerialization-aware checkpointing, towards the overall performance enhancement. Code implementation details can be found in Appendix~\ref{app:imple_details}.

\textbf{Cluster setup.} We evaluate our method and the baselines in (1) A single A100 DGX box with 8x80 GB GPUs. These GPUs are connected with NVLink; (2) 
2 DGX boxes with the same setting. These two boxes are interconnected by a stable 100 Gbps Infiniband.  This is a representative setting for cross-node training, where the communication overhead is large. Unless otherwise stated, this is our default setup. (3) Our in-house development cluster with 2x8 A100 40GB GPUs. This cluster has unstable inter-node bandwidth. Due to the limited computational budget, we report some peripheral results on this cluster.
% where conclusions can be drawn from a single-node setup or without involving cross-node training time.

% \DL{@Rulin Need your help on this motivation. ICLR reviewers have critized our models are curated. We probably want to show MHA and GQA as our main exps; and argue that 33H and less heads are also common patterns.}
\textbf{Model setup.} We evaluate our system on LLaMA-7B and its variants, encompassing four sets of model architectures in total: two with regular attention heads and two with irregular ones. We note both categories are important in real-world applications.%\\
% \todo{rulin: I feel it looks strange to divide as "regular" v.s. "irregular workload"}

\textbf{With regular attention heads.} 
% We focus on the original multi-head attention (MHA) and the recent popular grouped-query attention (GQA) architecture: 
(1) multi-head attention (MHA) models: LLaMA-7B with 4096 hidden size and 32 self-attention heads~\citep{touvron2023LLaMA}; (2) grouped-query attention(GQA) models: LLaMA-GQA~\citep{ainslie2023gqa}, same as LLaMA-7B but with 8 key-value heads, each shared by 4 queries as a group. %~\todo{awkward description}. 
During attention computation, it will first replicate to 32 heads to perform matrix multiplication with the correct shape. %\\
% To test the system performance for more general architecture, we also measure the following architecture.

\textbf{With irregular attention heads.} In addition, we benchmark the following variants that have appeared in applications but have not received much attention regarding their system efficiency: (3) models with an irregular (e.g., non-power-of-two) number of attention heads\footnote{For example, GPT-2-XL has 25 attention heads, GPT-2 has 12 attention heads, LLaMA-33B and its fine-tuned versions (e.g., Tulu-30B) have 52 attention heads, Whisper-large has 20 attention heads, and Falcon-7B has 71 attention heads~\citep{radford2019language, almazrouei2023falcon, ivison2023camels}.}: %a common source of inefficiency or vulnerability of contemporary distributed systems comes from the inflexibility to support irregular shape of input. For instance, embedding layer is often padded for a multiple of 128 for computing efficiency reason~\footnote{https://github.com/bigscience-workshop/Megatron-DeepSpeed/blob/main/megatron/arguments.py#L369}. 
%\DL{Shall we name these models? Reviewers may ask why don't we just benchamrk them.}\rulin{either we demonstrate our curated 33H setting is reasonable by listing these real-world examples, or we delete them and make a less convincing argument. Which one do you think is better? would be great if we have resources to run with one of these models instead}
%\DL{How about this: here we say A standard set of benchamrks to run in attention module is different number of attention heads (see flashattention tests), so we curate similar tests here to compare different systems performance. In discussion, we give examples why flashattention and we do these tests (See updated discussion). Basically, here our tone is we just say out this is a standard set of system benchmarks. Give any possible pressure to flashattention(I.e. any reviewers have to buy in what tri is doing) - if no one cares arbitrary heads, why tri dao bothers to spend so much time debugging and optimizing those heads? Then in discussion, we provide examples on real-world model for still not fully understood reviewers. Basically our role here is not to convince reviewers why this is correct, rather we provide examples to help reviewers understand why tri is doing all these.}
% In the context of the attention operation, many optimization efforts have been spent on efficiently arbitrary number of attention heads~\footnote{For instance, see a list of commits that FlashAttention makes for specific number of \href{https://github.com/Dao-AILab/flash-attention/blob/main/flash_attn/flash_attn_triton.py}{heads}.}. \rulin{removing this as this argument sounds weak.}
We intentionally test our systems and baselines on LLaMA-33H, which has the same configuration as LLaMA-7B but with 33 normal self-attention heads per layer. (4) models with fewer attention heads\footnote{~\citet{liu2021multi} finds fewer attention heads with more layers increase the performance.}: According to the recipe in \citet{liu2021multi}, we designed LLaMA-16H, LLaMA-8H, LLaMA-4H, and LLaMA-2H with 16, 8, 4, and 2 heads, respectively, as a proof of concept for situations when the number of heads is insufficient to further scale up model parallelism with limited resources. We keep the number of attention heads by scaling the number of layers properly\footnote{For instance, LLaMA-7B has 32 attention heads and 32 layers, thus LLaMA-16H has 16 attention heads per layers and 64 layers.} and keep the intermediate FFN layer size the same to make the model sizes still comparable. For example, LLaMA-16H has 16 attention heads per layer, a hidden size of 2048, an FFN layer of size 11008, and 64 layers.

% We keep the number of attention heads by scaling the number of layers properly according to~\citet{liu2021multi}, We also keep the intermediate FFN layer size the same to keep the number of parameters roughly the same so that the memory acquired by parameters is roughly the same.% 

\begin{table}[t]
    \centering
    \caption{Per iteration wall-clock time of~\attnname and Megatron-LM~\citep{korthikanti2023reducing} (Unit: seconds). Speedup in bold denotes the better of the two systems in the same configuration. Time measured with 2 DGX boxes.}
    \resizebox{.85\textwidth}{!}{
    \begin{tabular}{ccccccccccc}
        \toprule
        Method & \# GPUs & \multicolumn{2}{c}{Sequence Length} &  \multicolumn{2}{c}{LLaMA-7B}& \multicolumn{2}{c}{LLaMA-GQA} & \multicolumn{2}{c}{LLaMA-33H} \\ 
        & & Per GPU & Total & Time & speedup & Time & speedup & Time & speedup\\
        \midrule
        % \multirow{4}{*}{Megatron-LM} & 1x8 & 4K & 32K & 2.54 & 1.0x & 2.43 & 1.0x & 3.15 & 1.0x\\
        \multirow{3}{*}{Megatron-LM} & 1x8 & 8K & 64K & 6.81 & 1.0x & 6.60 & 1.0x & 8.37 & 1.0x\\
        & 1x8 & 16K & 128K & 20.93 & 1.0x & 20.53 & 1.0x & 25.75 & 1.0x\\ 
        & 1x8 & 32K & 256K & 72.75 & 1.0x & 71.93 & 1.0x & 90.21 & 1.0x\\
        \midrule
        % \multirow{4}{*}{\attnname} & 1x8 & 4K & 32K & 2.50 & \textbf{1.02x} & 2.30 & \textbf{1.06x} & 2.58 & \textbf{1.22x} \\
        \multirow{3}{*}{\attnname} & 1x8 & 8K & 64K & 5.98 & \textbf{1.14x} & 5.61 & \textbf{1.18x} & 6.08 & \textbf{1.38x}\\
        & 1x8 & 16K & 128K & 17.26 & \textbf{1.21x} & 16.86 & \textbf{1.22x} & 17.77 & \textbf{1.45x}\\
        & 1x8 & 32K & 256K & 58.46 & \textbf{1.24x} & 57.01 & \textbf{1.26x} & 59.96 & \textbf{1.50x}\\
        \midrule
        % \multirow{4}{*}{Megatron-LM} & 2x8 & 4K & 64K &  5.29 & 1.0x & 5.26 & 1.0x & 7.52 & 1.0x\\
        \multirow{3}{*}{Megatron-LM} & 2x8 & 8K & 128K & 14.26 & 1.0x & 14.21 & 1.0x & 20.63 & 1.0x\\
        & 2x8 & 16K & 256K & 43.44 & 1.0x & 43.20 & 1.0x & 62.78 & 1.0x\\
        & 2x8 & 32K & 512K & 147.06 & 1.0x & 146.38 & 1.0x& 216.70 & 1.0x\\
        \midrule
        % \multirow{4}{*}{\attnname} & 2x8 & 4K & 64K & 6.85 & 0.77x & 4.92 & \textbf{1.07x} & 7.03 & \textbf{1.07x}\\
        \multirow{3}{*}{\attnname}& 2x8 & 8K & 128K & 12.75 & \textbf{1.12x} & 9.74 & \textbf{1.46x}& 13.12 & \textbf{1.57x}\\
        & 2x8 & 16K & 256K & 30.21 & \textbf{1.44x} & 28.49 & \textbf{1.52x} & 31.33 & \textbf{2.00x}\\
        & 2x8 & 32K & 512K & 106.37 & \textbf{1.38x} & 102.34 & \textbf{1.43x} & 107.76 & \textbf{2.01x}\\
        \bottomrule
     \end{tabular}
     }
     \vspace{-0.3cm}
    \label{tab:main_results}
\end{table}

%\begin{table}[ht]
%    \centering
%    \begin{tabular}{cccccc}
%        \toprule
%        Seq. Per GPU & 4K & 8K & 16K & 32K \\
%        \midrule
%        \sysname &  150.69 & 300.61 & 784.35 & 2981.66 \\
%        Compute time & 52.34 & 185.62 & 722.30 & 2892.9 \\
%        Megaton-LM comm time & 
%        \bottomrule
%    \end{tabular}
%    \caption{A single Attention forward and backward in the cross node DGX cluster for communication time reference (Unit: ms).}
%    \label{tab:exp_micro}
%\end{table}

\begin{wraptable}{r}{0.5\textwidth}
    \centering
    \vspace{-3mm}
    \caption{The maximal sequence length Per GPU supported by~\attnname and Megatron-LM with tensor parallelism and pipeline parallelism on 16xA100 40GB GPUs.} %~\attnname supports 512K sequence length in all models, while Megatron-LM strategy maximal sequence length decreases with fewer heads, with either data parallelism or pipeline parallelism.}
    \resizebox{.5\textwidth}{!}{
    \begin{tabular}{cccccc}
        \toprule
        & 16H & 8H & 4H & 2H \\
        \midrule
        Megatron TP+DP & 512K & 256K & 128K & 64K 
        &\\
        Megatron TP+PP & 512K & 256K & 256K & 128K &\\
        ~\sysname  & 512K & 512K & 512K & 512K \\
        \bottomrule
    \end{tabular}
    }
    \vspace{-0.2cm}
    \label{tab:exp_scale}
\end{wraptable}

\subsection{Comparison with Megatron-LM on MHA and GQA models}
\label{sec:exp_faster}
%In this section, we compare our method with Megatron-LM on two settings: (1) the multi-head attention (MHA) models where the number of key and value heads equals the number of query heads, and (2) the grouped-query attention (GQA) models where the number of key and value heads is less than the number of query heads.% (3) the models with arbitrary numbers of heads, i.e. the number heads is unnecessarily a multiple of the parallelism degree.
% We show our method achieves faster training speed across almost all settings than Megatron-LM, when scaling to longer sequences.
% ~\DL{@Rulin, Seems now moved to the takeaway and setup section. Let's add some of the words here to there and remove this?}

\paragraph{Multi-head attention (MHA).}
On the LLaMA-7B model 
 (Table~\ref{tab:main_results}), our method achieves \textbf{1.24$\times$} and \textbf{1.44$\times$} speedup compared to Megatron-LM in single-node and cross-node setting, up to the longest sequence length we experiment. This is a joint result of our overlapping communication technique and our rematerialization-aware checkpointing strategy. %^. We take the configuration of 2x8 GPUs and 16K per GPU as an example. A single layer of LLaMA-7B takes 
We analyze how much each factor contributes to this result in the ablation study (\S~\ref{sec:ablation}).% and the performance on shorter sequences in \S~\ref{sec:discussion}. %\rulin{fix it%We do note that our method does not achieve better performance in shorter sequences, such as per GPU 4K setting for cross node. This is because the communication dominates the training run-time, where the overlapping technique can not reduce much (\S~\ref{sec:discussion}). %\rulin{fix it}.
%We leave the optimization of P2P communication on MHA models and shorter sequence length as an exciting future work.

\textbf{Grouped-query attention (GQA).}
On GQA model, \attnname communicates less volume due to the reduction of size of keys and values. On the contrary, the communication of Megatron-LM remains the same because it does not communicate keys and values. Thus, ~\attnname achieves a higher speedup on LLaMA-GQA model (Table~\ref{tab:main_results}).

\subsection{Comparison with Megatron-LM on models with irregular or less number of heads}%\rulin{directly say "scaling beyond head constraints" or "Comparison with Megatron-LM on models with XXX (less or irregular) heads"? "more general architecture" also includes other non-LLaMA model architectures}}
%\DL{agree, nice title!}
\label{sec:exp_megatron}
\paragraph{In support of irregular numbers of heads.}
Megatron-LM assumes the number of attention heads is divisible by the model parallelism degree. For example, it supports parallelism degrees of 2, 4, 8, 16, and 32 for models with 32 attention heads. However, it needs to pad dummy heads when the number of heads is not divisible by the ideal parallelism degree. For example, it needs to pad 15 dummy heads to support a parallelism degree of 16 for models with 33 attention heads (e.g., LlmaMA-33H), leading to a substantial computation wastage of 45.5\%. As shown in Table~\ref{tab:main_results}, we observe a \textbf{1.50$\times$} and \textbf{2.01$\times$} speedup (an additional 20\% and 45\% speedup compared to LLaMA-7B cases, aligned with the theoretical analysis).

% With LLaMA-33H models (Table~\ref{tab:main_results}), Megatron-LM exhibits an additional performance decline compared to~\sysname. This is due to its requirement to pad the number of attention heads so that the number of attention heads is divisible by the number of devices. On the other hand, ~\sysname does not need to partition attention heads and can support an arbitrary number of heads efficiently. For instance, when using 8 GPUs, Megatron-LM must pad the attention heads to 40, resulting in 21.2\% of the computation being wasted. In the case of 16 GPUs, Megatron-LM is compelled to pad the attention heads to 48, leading to a more substantial computation wastage of 45.5\%. 
% This roughly corresponds to a 1.21$\times$ or 1.45$\times$ increase in run-time compared to ~\sysname when training a LLaMA-7B model. This performance degradation of Megatron-LM is primarily because the training time is dominated by the attention module's computation time when scaling to longer sequence lengths. Empirically, we observe a \textbf{1.50$\times$} and \textbf{2.01$\times$} speedup (an additional 20\% and 45\% speedup compared to LLaMA-7B cases, aligned with the theoretical analysis).

\paragraph{In support of less number of heads.}
%Scaling beyond the number of heads.}
%Assuming the number of heads being a multiple of the tensor parallelism degree constraints Megatron-LM to scale its tensor parallelism degree beyond the number of heads, thus limiting its scaling ability to longer sequence lengths. 
When the number of GPUs exceeds the number of attention heads, Megatron-LM allows three possible solutions: (1) Pad dummy heads as in the LLaMA-33H scenario. However, the percentage of dummy heads almost directly translates to the percentage of slowdown in long sequences where attention computation dominates. (2) Use data parallelism for excess GPUs. %For instance, a user with 16 GPUs can choose to use 4-way data parallelism and 4-way tensor parallelism on the LLaMA-4H model. 
However, data parallelism does not reduce per sequence memory usage, and thus can not jointly support longer sequences.
%and thus the system can only support sequences as if the user only has 4 GPUs. 
(3) Use pipeline parallelism. However, the memory usage at each stage of the pipeline is not evenly distributed, limiting the maximal sequence length supported. For instance, in the LLaMA-2H experiment, we find that different stages consume from 18GB to 32GB in a 64K sequence length (Section~\ref{sec:mem_pp}). In addition, using pipeline parallelism introduces an extra fraction of GPU idle time. We demonstrate the effect of using the latter two solutions in Table~\ref{tab:exp_scale}. In 16 A100 40GB GPUs, \sysname supports 2$\times$ and 8$\times$ longer sequences.

%\dl{@anze, write this, why we use pipeline(because dp is an even worse choice), how we use it, why it can not scale; how we design model 8H, 4H, 2H. (1) We keep attention heads the same following jiawei's paper, (2) we keep the second ffn the same, to keep number of parameters the same.}

 % For example, Megatron-LM can only set a tensor model parallelism degree up to 8 if the model only has 8 attention heads in total.\DL{@rulin can you merge these sentences? they look great.}

\subsection{Comparison with Ring Self-Attention (RSA) and Ring Attention}
\label{sec:exp_rsa}
% \rulin{shall we move this to front as we are highlighting it as the key baseline in this version?}
% \DL{I think our main baseline remains Megatron-LM+FlashAttention. RSA is too old.}
Ring self-attention (RSA)~\citep{li2021sequence} communicates tensors in a ring fashion. We first report the maximal sequence length of RSA and~\sysname in Table \ref{tab:scale_rsa}, and found that~\sysname supports at least 8x longer sequences than RSA. This is mainly because RSA is not natively compatible with memory-efficient attention.
% (i.e. it modifies the attention computation\todo{rulin: I don't quite understand this i.e. so I removed it}). 
We further measure the iteration time with the maximal sequence length that RSA can support in Table~\ref{tab:speed_rsa}, and find that~\sysname is 4.45x - 5.64x faster than RSA. This speedup includes a 2x improvement from our causal workload balancing optimization and additional gains from the overlapping optimization and extending memory-efficient attention to the distributed setting. Both experiments are conducted with the Llama-7B model and on the DGX cluster.
% This is mainly because ~\sysname has optimized sequence parallelism in (1) causal language objective ($~$ 2x speedup), and (2) memory-efficient attention~\citep{dao2023flashattention}.  In explanation, memory-efficient attention also speeds up the attention computation by carefully managing the IO during attention, as pointed out in~\citep{dao2022flashattention, dao2023flashattention}. 
\begin{table}[ht]
    \centering
    \caption{Max sequence length and per iteration time (seconds) compared with RSA.}
    \begin{minipage}{.5\linewidth} 
      \centering
      \label{tab:scale_rsa}
      \resizebox{.9\textwidth}{!}{
      \begin{tabular}{ccc}
        \toprule
        & 1 Node & 2 Nodes\\
        \midrule
        RSA & 32K & 64K  \\
        \sysname & $>$ 256K & $>$ 512K\\
        \bottomrule
      \end{tabular}
      }
    \end{minipage}%
    \begin{minipage}{.5\linewidth} 
      \centering
      \label{tab:speed_rsa}
      \resizebox{.9\textwidth}{!}{
      \begin{tabular}{ccc}
        \toprule
        & 1 Node (32K) & 2 Nodes (64K)\\
        \midrule
        RSA & 14.10 & 30.49 \\
        \sysname & 2.50 & 6.85\\
        \midrule
        Speedup & 5.64x & 4.45x \\
        \bottomrule
      \end{tabular}
      }
    \end{minipage}
    %\vspace{-5mm}
\end{table}

Ring Attention~\citep{liu2023ring} implements distributed attention in a memory-efficient manner. The key difference between Ring Attention and \sysname is \sysname has additional optimization of causal workload balancing and a better gradient checkpoint strategy.  The implementation of Ring Attention uses a different framework from ours (Jax versus PyTorch). To provide a fair comparison, we consider our ablation version in \S~\ref{sec:ablation} as a PyTorch implementation of Ring Attention. 
% To provide a fair comparison, we compare with its design in PyTorch, which is \sysname without load balancing (\S~\ref{sec:scheduling}). 
\S~\ref{sec:ablation} provides a detailed analysis. In 8-GPU setting, we observe a 1.67$\times$ speedup (7.5$\times$ versus 4.5$\times$ speedup compared to a single GPU FlashAttention) over the design of Ring Attention.

\subsection{Comparison with DeepSpeed Ulysses}
\label{sec:comp_ds}
\begin{table}[ht]
    \centering
    \caption{Per iteration wall-clock time (seconds) of~\sysname and DeepSpeed Ulysses.}
    \resizebox{\linewidth}{!}{
    \begin{tabular}{c|ccccc|ccccc}
        \toprule
        Method & \# GPUs & \multicolumn{2}{c}{Sequence Length} &  Time & Speedup & \# GPUs & \multicolumn{2}{c}{Sequence Length} &  Time & Speedup \\ 
        & & Per GPU & Total & & & & Per GPU & Total & & \\
        \midrule
         & \multicolumn{5}{c}{Llama-7B} & \multicolumn{5}{c}{Llama-33H} \\
        \midrule
        % \multirow{4}{*}{DeepSpeed-Ulysses} & 2x8 & 4K & 64K & 4.29 &  \textbf{1.23x} & 2x8 & 4K & 64K & 6.42 &  \textbf{1.17x}\\
        % & 2x8 & 8K & 128K & 11.61 & \textbf{1.23x} & 2x8 & 8K & 128K & 17.47 & 1.18x \\
        \multirow{2}{*}{DeepSpeed-Ulysses}& 2x8 & 16K & 256K & 37.53 & 1.0x & 2x8 & 16K & 256K & 56.63 & 1.0x\\
        & 2x8 & 32K & 512K & 134.09 & 1.0x & 2x8 & 32K & 512K & 202.89 & 1.0x \\
        \midrule
        % \multirow{4}{*}{\sysname} & 2x8 & 4K & 64K & 6.85 & 0.77x & 2x8 & 4K & 64K & 7.03 & 1.07x\\
        % & 2x8 & 8K & 128K & 12.75 & 1.12x & 2x8 & 8K & 128K & 13.12 & \textbf{1.57x} \\
        \multirow{2}{*}{\sysname}& 2x8 & 16K & 256K & 30.21 & \textbf{1.21x} & 2x8 & 16K & 256K & 31.33 & \textbf{1.81x} \\
        & 2x8 & 32K & 512K & 106.37 & \textbf{1.26x} & 2x8 & 32K & 512K & 107.76 & \textbf{1.88x}\\
        \bottomrule
     \end{tabular}%}
    }
     %Time measured with 2 DGX boxes.}
    \label{tab:comp_ds}
    %\vspace{-5mm}
\end{table}

DeepSpeed-Ulysses~\citep{jacobs2023deepspeed} uses all-to-all primitive to reduce the communication. We evaluate a representative subset of experiments in Table~\ref{tab:comp_ds} due to computational budget limit. On experiments with regular heads models (Llama-7B), ~\sysname achieves 1.26$~\times$ speedup. On experiments on irregular heads models (Llama-33H), ~\sysname achieves 1.88$\times$ speedup. Essentially, DeepSpeed-Ulysses also paralleize on the attention head dimension, and suffer from the same problems as analyzed in~\S~\ref{sec:exp_megatron}. %We further summarize a discussion on whether future systems should parallelize on attention heads or sequence dimension in Appendix~\ref{sec:discussion}.

% Yet, as it is also partitioning the attention head dimension, it suffers from similar problems as analyzed above. %We provide some end-to-end comparisons in this section. %We note that the communication in DeepSpeed Ulysses can be faster than \sysname, especially with shorter context length and slower network, where the overlapping technique in \sysname cannot perfectly hide all the communication. This can be potentially addressed by optimizing the P2P communication as discussed above.
%We run a subset of the experiments compared with DeepSpeed-Ulysses (Table~\ref{tab:comp_ds}). %Firstly, DeepSpeed-Ulysses does reduce the communication overhead, and thus better than Megatron-LM on scenarios listed in Table~\ref{tab:comp_ds}. 
%~\sysname achieves better performance than DeepSpeed-Ulysses on long sequences or models with a more general number of heads (e.g. Llama-33H). We also note that DeepSpeed-Ulysses can not scale beyond the number of attention heads because it also relies on sharding the attention heads. However, we need to point out that in shorter sequences and MHA models (where ~\sysname does not have a communication advantage, compared to GQA/MQA models), the communication primitives used in DeepSpeed-Ulysses are more advantageous. We leave our further optimization in P2P in shorter sequences and MHA models as an exciting future work.

\subsection{Ablation Study}
\label{sec:ablation}
\begin{figure}[t]
    \centering
    \begin{minipage}{0.33\textwidth}
        \centering
        \includegraphics[width=\textwidth]
        {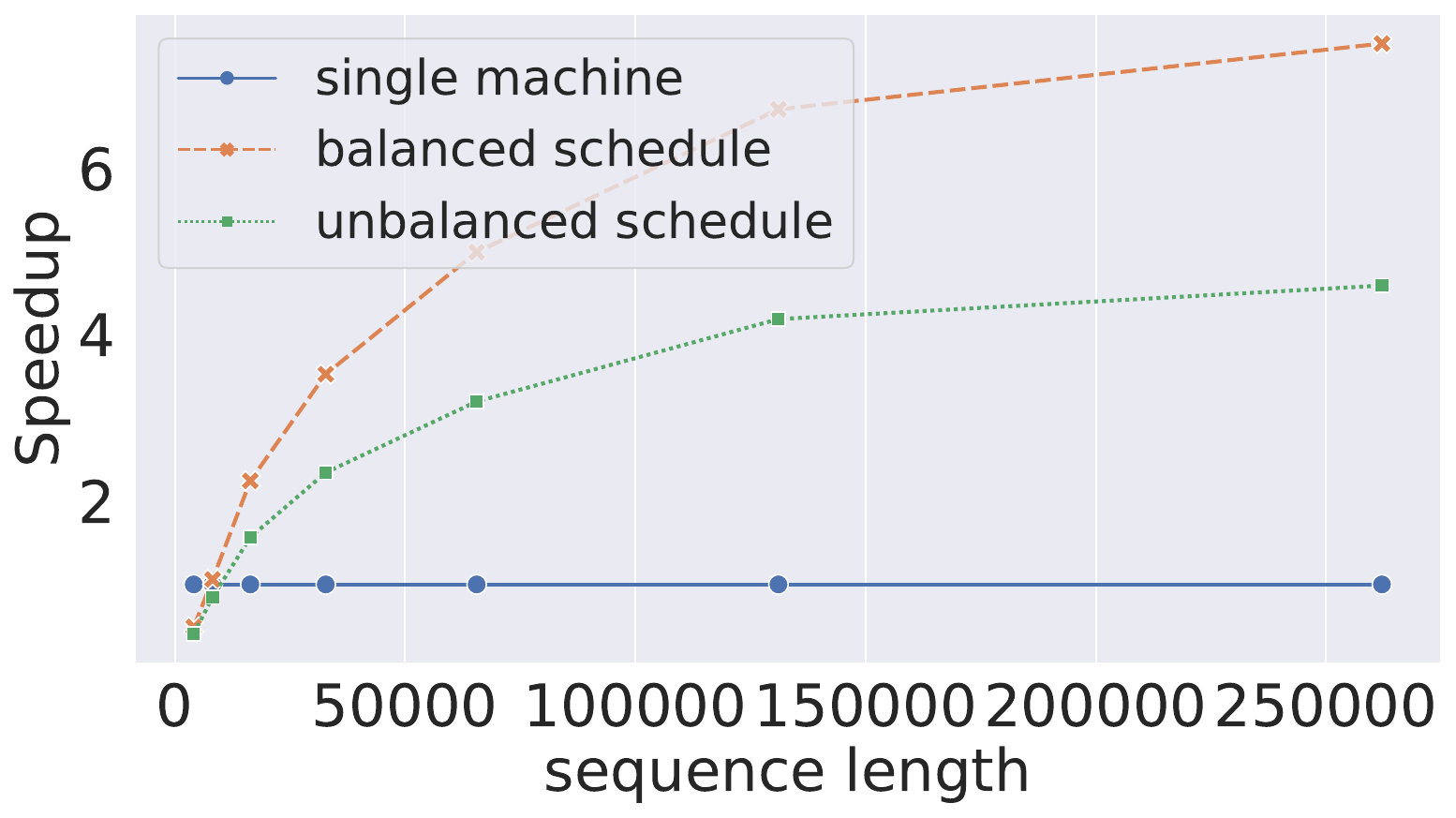}
    \end{minipage}\hfill % add some space between figures if needed
    \begin{minipage}{0.66\textwidth}
        \centering
        \includegraphics[width=\textwidth]
        {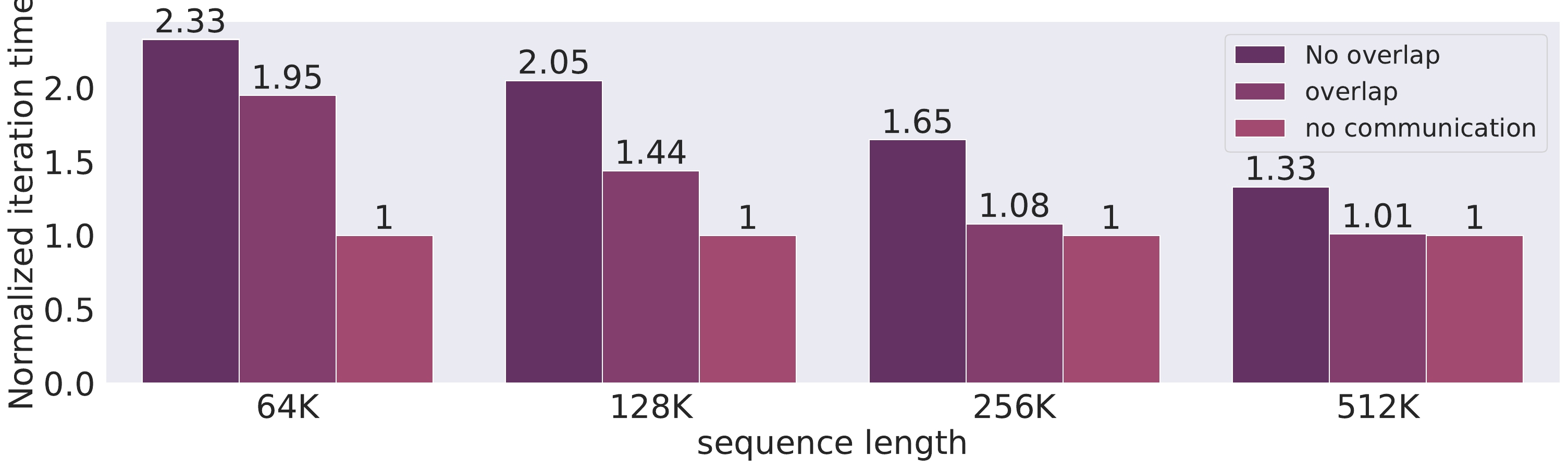}
    \end{minipage}
    \caption{Effect of balanced schedule (left) and the effect of overlapping (right).}
    \label{abl:load_overlap}
    \vspace{-2mm}
\end{figure}

%In this section, we ablate the gain of each system optimizations in~\sysname.
\paragraph{Effect of Load Balancing} We study load balancing on an attention forward pass of LLaMA-7B model, on 8 A100 40GB GPUs (Figure~\ref{abl:load_overlap}). The backward pass follows a similar analysis. With an unbalanced schedule (Figure ~\ref{fig:load_balance}), the total work done is 36, where the total work could be done in 8 units of time is 64. Thus, the expected speedup is 4.5x. In the balanced schedule, the expected speedup is 7.2x. We scale the total sequence length from 4K to 256K. The unbalanced version saturates in 4.5x speedup compared to a single GPU FlashAttention, while the balanced version saturates at 7.5$\times$ %~\footnote{The observed higher speedup is because FlashAttention kernel flops drop for longer sequences.} 
speedup. Both of them align with our theoretical analysis and show the effectiveness of the balanced scheduling. 

%\begin{figure}
%    \centering
%    \includegraphics[width=0.45\textwidth]{figures/load_balance_result.pdf}
%    \vspace{-0.4cm}
%    \caption{Ablation on the effect of balanced schedule.}
%    \label{abl:load_balance}
%\end{figure}
%\begin{figure}
%    \centering
%    \includegraphics[width=0.45\textwidth]{figures/abl_overlap.pdf}
%    \vspace{-0.4cm}
%    \caption{Ablation on the effect of overlapping.}
%    \label{abl:overlap}
%\end{figure}

\begin{wraptable}{r}{0.5\textwidth}
    \centering
    \vspace{-0.5cm}
    \caption{Our checkpointing algorithm (``Our ckpt'') versus HuggingFace strategy (``HF ckpt'') on 8 A100s (batch size 1, Unit: seconds).}
    \resizebox{0.5\columnwidth}{!}{\begin{tabular}{ccccccc}
        \toprule
        Method & \multicolumn{6}{c}{ Sequence Length Per GPU} \\
        & 1K & 2K & 4K & 8K & 16K & 32K \\
        \midrule
        HF ckpt & 0.84 & 1.29 & 2.64 & 6.93 & 21.44 & 76.38 \\
        Our ckpt & 0.84 & 1.36 & 2.50 & 5.98 & 17.26 & 58.46 \\
        \midrule
        Speedup & 1.0x & 0.94x & \textbf{1.06x} & \textbf{1.16x} & \textbf{1.24x} & \textbf{1.31x} \\
        \bottomrule
    \end{tabular}}
    \vspace{-0.2cm}
    \label{tab:abl-ckpt}
\end{wraptable}

\paragraph{Effect of overlapping communication and computation.}
We study the overlapping communication on LLaMA-7B and 2 DGX boxes (Figure~\ref{abl:load_overlap}). We find that overlapping greatly reduces the communication overhead. On a global sequence length of 128K, the communication overhead is reduced from 105\% to 44\%. This overlapping scheme maximizes its functionality when the communication overhead is less than 100\%, where all communication can be potentially overlapped. Empirically, we find the system only exhibits 8\% and 1\% overhead in these cases, a close performance to an ideal system without communication.

\paragraph{Effect of rematerialization-aware checkpointing.} 
We show in Table~\ref{tab:abl-ckpt} the effects of the proposed rematerialization-aware gradient checkpointing. Our method achieves 1.16x, 1.24x, and 1.31x speedup at the sequence length of 8K, 16K, and 32K per GPU respectively. The materialization-aware checkpointing strategy speeds up more at longer sequence lengths where the attention dominates the computation.

\subsection{Partition on the attention heads or sequence dimension}
\label{sec:discussion}
%We summarize thoughts from comparison with Megatron-LM in the previous section. Then we discuss the future directions that can further improve \sysname. 
% We then compare our method with one concurrent open-sourced project which also splits the attention heads. Finally, we discuss the role of pipeline parallelism in supporting long sequence training and shows it is less effective than tensor parallelism, which is the reason we do not consider it as a major baseline.
%\paragraph{Partition on the attention heads or sequence dimension}
Megatron-LM and DeepSpeed-Ulysses are distributed systems that partition on attention heads. While it allows seamless integration with the FlashAttention kernel, it has certain limitations. These includes: (1) Not being able to utilize the pattern inside the attention module, missing opportunities to reduce communication for causal, and grouped-query attention (See \S~\ref{sec:comm_mem_analysis}). (2) not flexible to support arbitrary number of attention heads, % limiting an efficient support for a broader family of models, such as GPT-2-XL (25 attention heads), GPT-2 (12 heads), LLaMA-33B and its fine-tuned versions (e.g., Tulu-30B, 52 heads), Whisper-large (20 heads), and Falcon-7B (71 heads); 
and (3) Importantly, its scalability is limited by the number of attention heads (in the scale of several to several dozens), while the maximal number of parallelism degree for sequence parallelism is at least several thousands. Given these reasons, we think it is worth pursuing the sequence parallelism paradigm when distributing the attention module. 

\section{Conclusion}
In this work, we introduce~\sysname, a distributed memory-efficient attention prototype for long-context transformer training based on sequence parallelism. \sysname presents novel system optimizations including load balancing for causal language modelings, overlapped communication with computation in the distributed attention computation, and a re-materialization-aware checkpointing strategy. Experiments evaluate multiple families of transformer models and on different cluster types, and over four strong distributed system baselines. In particular, \sysname has demonstrated up to 2.01$\times$ speedup and scales up to 8x longer sequences, compared to the popular system, Megatron-LM with FlashAttention. %Future directions include implementing topology-aware P2P operations to further reduce training time in lower sequence lengths.
\newpage

%%%%%%%%%%%%%%%%%%%%%%%%%%%%%%%%%%%%%%%%%%%%%%%%%%%%%%%%%%%%%%%%%%%%%%%%%%%%%%%
%%%%%%%%%%%%%%%%%%%%%%%%%%%%%%%%%%%%%%%%%%%%%%%%%%%%%%%%%%%%%%%%%%%%%%%%%%%%%%%
% APPENDIX
%%%%%%%%%%%%%%%%%%%%%%%%%%%%%%%%%%%%%%%%%%%%%%%%%%%%%%%%%%%%%%%%%%%%%%%%%%%%%%%
%%%%%%%%%%%%%%%%%%%%%%%%%%%%%%%%%%%%%%%%%%%%%%%%%%%%%%%%%%%%%%%%%%%%%%%%%%%%%%%

\bibliography{colm2024_conference}
\bibliographystyle{colm2024_conference}

\newpage
\appendix

\section{From FlashAttention to ${attn}(\cdot)$ in \sysname}
\label{sec:attn_in_lightseq}

In this section, we provide the details of the $(attn)(\cdot)$ kernel in \attnname.% and how it can be used with the outer \sysname logic of the forward pass 
(Alg~\ref{alg:lightseq_fwd}). For conceptual simplicity, we demonstrate it in the most vanilla version, without the actual scheduling (e.g. load balancing and overlapping). We also demonstrate it with the causal language modeling objective. The standalone attention is mainly borrowed from the FlashAttention2 paper~\citep{dao2023flashattention}. To make it compatible with \attnname, we mainly revised the several points:
\begin{enumerate}
    \item Accumulate results statistics $o$, $m$ and $l$ from previous computation, instead of initializing them inside the function.
    \item Pass an extra argument "last", which means whether this is the last chunk of attention computation. Only when it is true, we compute the logsumexp $L$.
\end{enumerate}
At a high level, on a worker $p$, \sysname first initializes local statistics $o, m, l, L$. Then \sysname loops over all its previous workers. In each iteration, it fetches the key and the value from a worker and invokes the revised standalone attention to update local statistics. At the end of the iteration, it needs to delete the remote key and value from HBM so that the memory does not accumulate. At the last iteration of the loop, it additionally calculates the logsumexp according to the final $m$ and $l$ (triggered by the "last" variable in the algorithm).  At the end of the forward pass, worker $p$ has the correct $m, l, L$. The backward pass is similar and conceptually simpler because we do not need to keep track of statistics such as $m$ and $l$. Instead, we only need to use the logsumexp stored in the forward pass. 

\begin{algorithm}
  \caption{(Vanilla)~\attnname of worker $p$ \small\label{alg:distflashattn_forward_highlevel_unbalanced}}
  \begin{algorithmic}[1]
    \REQUIRE $\mathbf{q}_p, \mathbf{k}_p, \mathbf{v}_p$ 
    \STATE Initialize $\mathbf{o}_p$ = $\mathbf{o}^0$, $\mathbf{s}_p =\mathbf{s}^0 = [\mathbf{m}^0, \mathbf{l}^0]$, where $\mathbf{o}^0=\mathbf{0}$, $\mathbf{l}^0=\mathbf{0}$, and $\mathbf{m}^0=[-\mathbf{\infty} \cdots -\mathbf{\infty}]^T$\\
    \STATE $\mathbf{o}_{p}$, $\mathbf{s}_{p}$ = ${attn}(\mathbf{q}_p, \mathbf{k}_{p}, \mathbf{v}_{p}, \mathbf{o}_p, \mathbf{s}_p)$ \\
    \FOR{{$1 \le t < p$}}
          \STATE r = $(p-t) \pmod P$
          \STATE Fetch from remote: worker p $\xleftarrow{\mathbf{k}_r, \mathbf{v}_r}$ worker r
          \STATE $\mathbf{o}_{p}$, $\mathbf{s}_{p}$ = ${attn}(\mathbf{q}_p, \mathbf{k}_{r}, \mathbf{v}_{r}, \mathbf{o}_{p}, \mathbf{s}_{p})$ \\
    \ENDFOR
    \STATE Return $\mathbf{o}_{p}$.
  \end{algorithmic}
\end{algorithm}

\begin{algorithm}
  \caption{(Balanced)~\attnname of worker $p$ \small\label{alg:distflashattn_forward_highlevel_balanced}}
  \begin{algorithmic}[1]
    \REQUIRE $\mathbf{q}_p, \mathbf{k}_p, \mathbf{v}_p$ \\
    \STATE Initialize $\mathbf{o}_p$ = $\mathbf{o}^0$ , $\mathbf{s}_p =\mathbf{s}^0 = [\mathbf{m}^0, \mathbf{l}^0]$, where $\mathbf{o}^0=\mathbf{0}$, $\mathbf{l}^0=\mathbf{0}$, and $\mathbf{m}^0=[-\mathbf{\infty} \cdots -\mathbf{\infty}]^T$\\
    \STATE $\mathbf{o}_{p}$, $\mathbf{s}_{p}$ = ${attn}(\mathbf{q}_p, \mathbf{k}_{p}, \mathbf{v}_{p}, \mathbf{o}_{p}, \mathbf{s}_{p})$ \\
    \FOR{1 $\le$ t $\le$ $\lfloor \frac{P}{2} \rfloor$ }
    \STATE $r = (p-t) \pmod P$\\
          \IF{$p$ $>$ t}
          \STATE Fetch key, value from remote: p
          $\xleftarrow{\mathbf{k}_t, \mathbf{v}_t}$ r 
          \STATE $\mathbf{o}_{p}$, $\mathbf{s}_{p}$ = ${attn}(\mathbf{q}_p, \mathbf{k}_{r}, \mathbf{v}_{r}, \mathbf{o}_{p}, \mathbf{s}_{p})$ \\
          % \STATE r = $(p-t) \pmod P$
          \IF{t $\neq$ $\lfloor \frac{P}{2} \rfloor$ \AND $(p+t)$ $>$ P} 
          \STATE $r_2 = (p+t) \pmod P$ \\
          \STATE Fetch result from remote: p $\xleftarrow{\mathbf{o}_{p}^{'}, \mathbf{s}_{p}^{'}} $ $r_2$ \\
          \STATE $\mathbf{o}_{p}$, $\mathbf{s}_{p}$ = ${rescale}(\mathbf{o}_{p}, \mathbf{s}_{p}, \mathbf{o}_{p}^{'}, \mathbf{s}_{p}^{'})$ 
          \ENDIF
          \ELSE%{$p$ $\le$ t}
          % \STATE $r = P+1-t$ \\
                %\IF{$r>(\lfloor \frac{P}{2} \rfloor+1)$}
                \IF{t $\neq$ $\lfloor \frac{P}{2} \rfloor$}
                \STATE Fetch query from remote: p $\xleftarrow{\mathbf{q}_r} $ r 
                \STATE $\mathbf{o}_{r}$, $\mathbf{s}_{r}$ = ${attn}(\mathbf{q}_{r}, \mathbf{k}_{p}, \mathbf{v}_{p}, \mathbf{o}^0, \mathbf{s}^0)$
                \STATE Send result to remote: p $\xrightarrow{\mathbf{o}_{r}, \mathbf{l}_{r}, \mathbf{m}_{r}} $ r  
               % \ELSE
               % \STATE WAIT
                \ENDIF
          \ENDIF
          \\
    \ENDFOR
    \STATE Return $\mathbf{o}_{p}$.
  \end{algorithmic}
\end{algorithm}

\begin{algorithm}
  \caption{\small\label{alg:lightseq_fwd} \sysname Pseudo code (forward pass)}
  \begin{algorithmic}[1]
    \REQUIRE Matrices $\vQ^p, \vK^p, \vV^p \in \mathbb{R}^{\frac{N}{\mathbb{P}} \times d}$ in HBM, block sizes $B_c$, $B_r$, rank\\
    %\Function
    \textbf{function} {$\text{standalone\_fwd}$}{q, k, v, o, $\ell$, m, causal, last}
    % \STATE Divide $q$ into $T_r = \left\lceil\frac{N}{\mathbb{P}B_r} \right\rceil$ blocks $\q_1, \dots, \q_{T_r}$ of size $B_r \times d$ each,
    \STATE Divide $q$ into $T_r = \left\lceil\frac{N}{\mathbb{P}B_r} \right\rceil$ blocks $q_1, \dots, q_{T_r}$ of size $B_r \times d$ each,
    \STATE and divide $k, v$ in to $T_c = \left\lceil \frac{N}{\mathbb{P}B_c} \right\rceil$ blocks $k_1, \dots, k_{T_c}$ and
    $v_1, \dots, v_{T_c}$, of size $B_c \times d$ each.
    \STATE Divide the output $o \in \mathbb{R}^{\frac{N}{\mathbb{P}} \times d}$ into $T_r$ blocks $o_i, \dots, o_{T_r}$ of size
    $B_r \times d$ each, and divide the logsumexp $L$ into $T_r$ blocks $L_i, \dots, L_{T_r}$ of size
    $B_r$ each.
    \FOR{$1 \le i \le T_r$} \label{alg:standalone_attn_outer_loop}
      \STATE \label{alg:standalone_attn_load_q} Load $q_i$ from HBM to on-chip SRAM.
      \STATE \label{alg:standalone_attn_init} Load $o_i \in \mathbb{R}^{B_r \times d}$, $\ell_i \in \mathbb{R}^{B_r}$, $m_{i} \in \mathbb{R}^{B_r}$ from HBM to on-chip SRAM as $o_i^{(0)}$, $\ell_{i}^{(0)}$, $m_i^{(0)}$.
      %{\color{red} Load $o_i \in \mathbb{R}^{B_r \times d}$, $\ell_i \in \mathbb{R}^{B_r}$, $m_{i} \in \mathbb{R}^{B_r}$ from HBM to on-chip SRAM as $o_i^{(0)}$, $\ell_{i}^{(0)}$, $m_i^{(0)}$}
      %\STATE \label{alg:standalone_attn_init} On chip, initialize $\vO_{i}^{(0)} = (0)_{B_r \times d} \in \mathbb{R}^{B_r \times d}, \ell_{i}^{(0)} = (0)_{B_r} \in \mathbb{R}^{B_r}, m_{i}^{(0)} = (-\infty)_{B_r} \in \mathbb{R}^{B_r}$.
      \FOR{$1 \le j \le T_c$}
        \IF{causal and $i \le j$}
        \STATE Continue %\DL{Improve causal code, fa does in more fine-grained.}
        \ENDIF
        \STATE \label{alg:standalone_attn_load_kv} Load $k_j, v_j$ from HBM to on-chip SRAM.
        \STATE \label{alg:dist_attn_qk} On chip, compute $s_{i}^{(j)} = q_i k^{T}_j \in \mathbb{R}^{B_r \times B_c}$.
        \STATE \label{alg:standalone_attn_statistics} On chip, compute
        $m_{i}^{(j)} = \mathrm{max}(m_{i}^{(j-1)}, \mathrm{rowmax}(s_{i}^{(j)})) \in \mathbb{R}^{B_r}$, $\tilde{p}_{i}^{(j)} = \exp(S_{i}^{(j)} - m_{i}^{(j)}) \in \mathbb{R}^{B_r \times B_c}$ (pointwise),
        $\ell_{i}^{(j)} = e^{m_{i}^{j-1} - m_{i}^{(j)}} \ell_{i}^{(j-1)} + \mathrm{row sum}(\tilde{p}_{i}^{(j)}) \in \mathbb{R}^{B_r}$.
        \STATE \label{alg:standalone_attn_update} On chip, compute
        $o_{i}^{(j)} = \diag(e^{m_{i}^{(j-1)} - m_{i}^{(j)}})^{-1} o_{i}^{(j-1)} + \tilde{p}_{i}^{(j)} v^p_j$.
      \ENDFOR
      \STATE On chip, compute $o_{i} = \diag(\ell_{i}^{(T_c)})^{-1} o_{i}^{(T_c)}$.
      \STATE Write $o_{i}$ to HBM as the $i$-th block of $o$.
      \IF{last}
           \STATE On chip, compute $L_{i} = m_{i}^{(T_c)} + \log(\ell_i^{(T_c)})$.
           \STATE Write $L_{i}$ to HBM as the $i$-th block of $L$.
      \ENDIF
    \ENDFOR
    \STATE Return $o, \ell, m$ and the logsumexp $L$.
    % \textbf{end function}\\
    %\EndFunction
    \\
    \textbf{end function}
    %\STATE On another stream: preGroup nccl.send() on $K_p, V_p$ to all other processes.
    %\DL{can pre-fetch later}
    \STATE Initialize $\vO^p = (0)_{\frac{N}{\mathbb{P}} \times d} \in \mathbb{R}^{\frac{N}{\mathbb{P}} \times d}, \ell^{(p)} = (0)_{\frac{N}{\mathbb{P}}} \in \mathbb{R}^{\frac{N}{\mathbb{P}}}, m^p = (-\infty)_{\frac{N}{\mathbb{P}}} \in \mathbb{R}^{\frac{N}{\mathbb{P}}}$.
    \STATE $\vO^p$, $\ell^p$, $m^p$, $L^p$ = standalone\_fwd($\vQ^p, \vK^p, \vV^p$, $\vO^p$, $\ell^p$, $m^p$, True, p=1)
    %\FOR{{\color{red} $1 \le r < p$ \DL{causal}}} 
    \FOR{$1 \le r < p$}
        %\STATE r = 
        \STATE \label{alg:dist_attn_fetch} Receive $\vK^r$ and $\vV^r$ from \textbf{Remote} worker $r$ into HBM.
        \STATE $\vO^p$, $\ell^p$, $m^p$, $L^p$ = standalone\_fwd($\vQ^p, \vK^y, \vV^y$, 
        $\vO^p$, $\ell^p$, $m^p$, False, r=(p-1)
        \STATE Delete $\vK^r$ and $\vV^r$ from HBM.
    \ENDFOR
    \STATE Return the output $\vO^p$ and the logsumexp $L$.
  \end{algorithmic}
\end{algorithm}

\section{Load-balancing Algorithm for Causal Modeling}\label{app:load_balance}
\begin{figure*}[t!]
    \centering
    \includegraphics[width=0.7\linewidth]{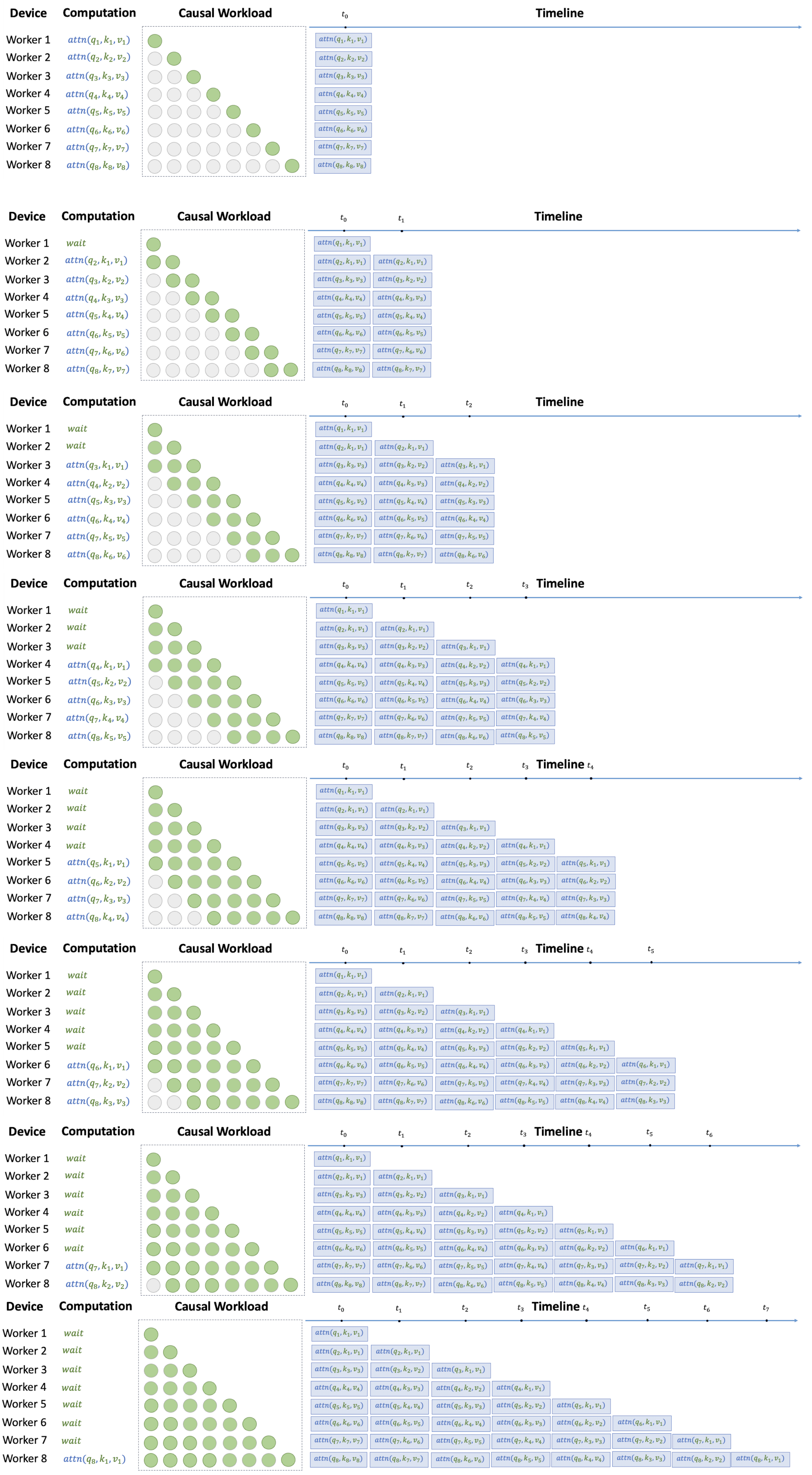}
    \caption{Illustration of \attnname before applying load-balancing on 8 workers.}
    \label{fig:before_balance}
\end{figure*}

\begin{figure*}[t!]
    \centering
    \includegraphics[width=0.7\linewidth]{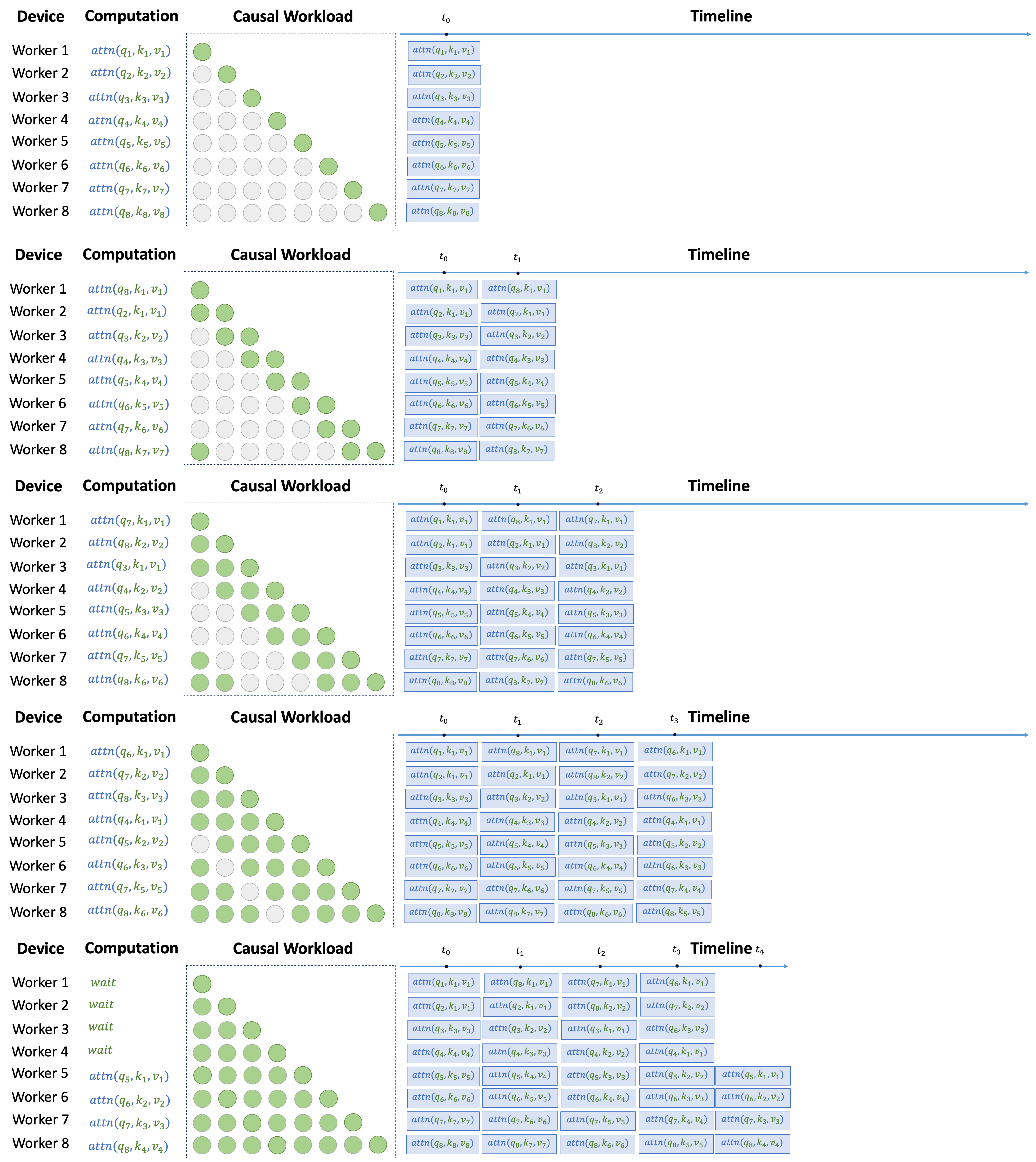}
    \caption{Illustration of \attnname after applying load-balancing on 8 workers.}
    \label{fig:after_balance}
\end{figure*}

In this section, we detail the design of our load-balancing algorithm for causal modeling. We show the workload of each worker in all time steps in Figure~\ref{fig:before_balance} (before applying load-balancing) and Figure~\ref{fig:after_balance} (after applying load-balancing) in an 8-worker scenario. The communication schema is also reflected in both figures by comparing the tensors each worker holds at the consecutive two time steps.

\section{Memory Consumption for Pipeline Parallelism}
\label{sec:mem_pp}
In this section, we show the memory consumption of Megatron-LM when training with tensor parallelism and pipeline parallelism. As presented in table \ref{tab:mem_consumption}, memory consumption are uneven across different pipeline stages, making scaling through pipeline parallelism hard. % Such imbalanced memory consumption is a major factor that hinders Megatron-LM to scale to longer sequence length through pipeline parallelism.

\begin{figure}[ht]
\centering
\includegraphics[width=0.45\textwidth]{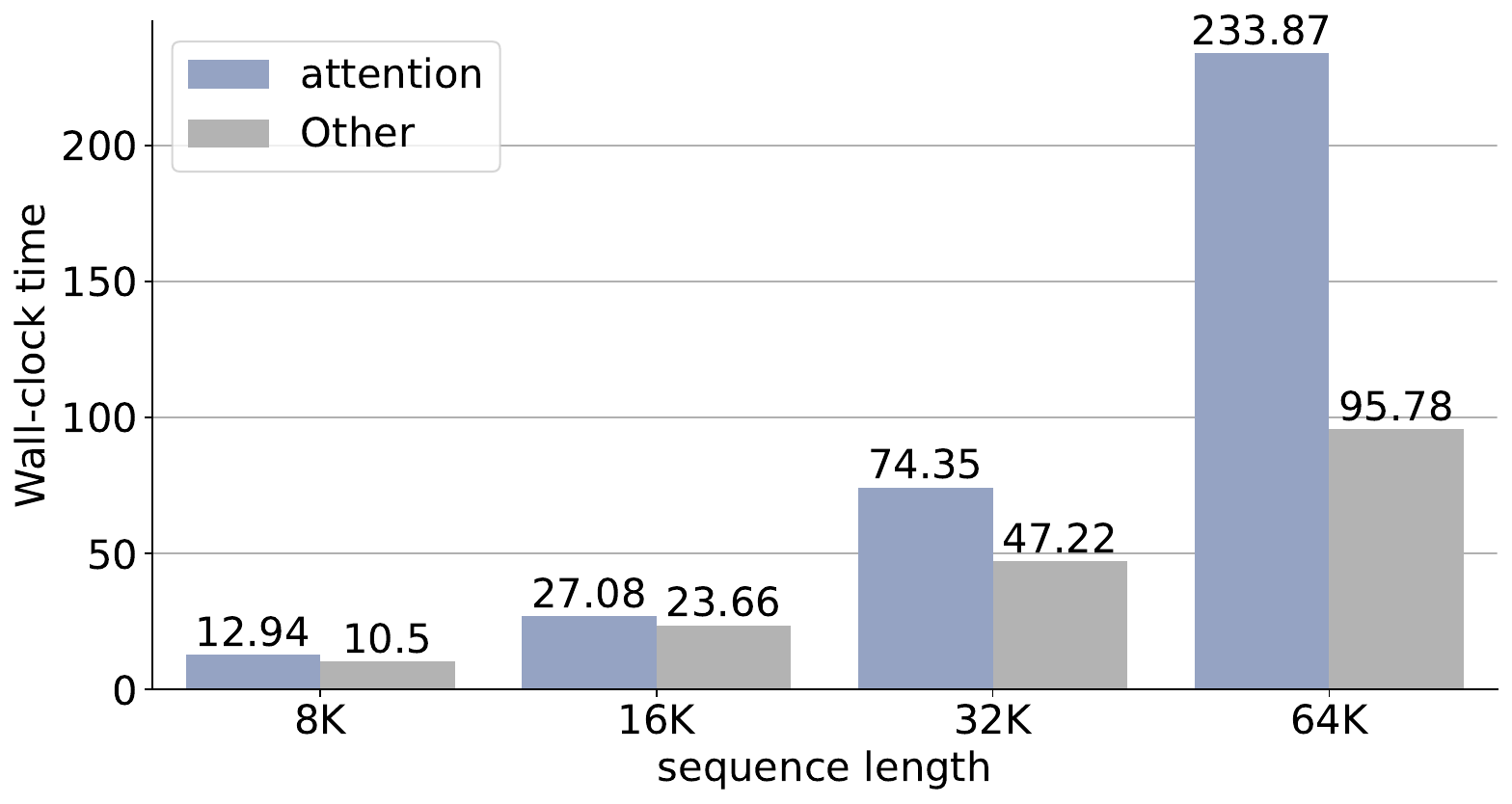}
\caption{Time breakdown of attention versus other modules in a forward pass, measured with Flash-Attention~\citep{dao2023flashattention} on a single 40GB A100 GPU. (Unit ms)} 
\label{fig:attn_time}
\end{figure}

\begin{table}[ht]
    \centering
    \caption{The memory consumption of Megatron-LM when training Llama-2H with tensor parallelism (degree=2) and pipeline parallelism (degree=8) on 16xA100 40GB GPUs at the sequence length of 128K. The memory consumption is highly uneven across pipeline stages.}
    \resizebox{0.95\linewidth}{!}{
    \begin{tabular}{cccccccccc}
        \toprule
        & Worker 1 & Worker 2 & Worker 3 & Worker 4 & Worker 5 & Worker 6 & Worker 7 & Worker 8  \\
        \midrule
        node 1 & 31.5GB & 31.4GB & 28.7GB & 28.7GB & 26.0GB & 26.0GB & 24.6GB & 24.6GB\\ 
        node 2 & 21.8GB & 21.8GB & 20.5GB & 20.5GB & 17.9GB & 17.8GB & 32.0GB & 32.1GB  \\
        \bottomrule
    \end{tabular}}
    \label{tab:mem_consumption}
\end{table}

\section{Communication and memory analysis}
\label{sec:comm_mem_analysis}
Denote the hidden dimension as $d$. In \attnname, every worker needs to fetch key and value chunks both of size $\frac{N}{P}d$ before performing the corresponding chunk-wise computation. Thus, the total communication volume in the $P$-workers system is $2\times \frac{N}{P}d \times P = 2Nd$. With the causal language objective, half of the keys and values do not need to be attended, halving the forward communication volume to $Nd$. In the backward pass, \attnname needs to communicate keys, values, and their gradients, which has $2Nd$ volume. It adds up to $3Nd$ as the total communication volume for \attnname. In Megatron-LM~\citep{korthikanti2023reducing}, each worker needs to perform six all-gather and four reduce-scatter on a $\frac{N}{P}d$ size tensor, thus giving a total communication volume of $10Nd$. Considering gradient check-pointing, Megatron-LM will perform communication in the forward again, giving a total volume of $14Nd$. On the other hand, our communication volume remains $3Nd$ because of the rematerialization-aware strategy. In conclusion, \sysname achieves 4.7x communication volume reduction compared with Megatron-LM. 

In large model training, we usually utilize techniques such as FSDP to also reduce the memory consumed by model weights. In this case, We note that the communication introduced by FSDP is only proportional to the size of model weights, which does not scale up with long sequence length. We show the end-to-end speedup with FSDP in Table~\ref{tab:main_results}. For clarity, we also note that~\sysname is orthogonal to FSDP and by default can be used by itself.
%In practice, we combine \sysname with FSDP to also distribute the model weights for large models. We note that the communication introduced by FSDP is only proportional to the size of model weights, which does not scale up with long sequence length. We show the end-to-end speedup with FSDP in Table~\ref{tab:main_results}.
In the situations where the model uses MQA or GQA, \sysname further saves the communication volumes by the shared key and values, which we discuss in detail in \S~\ref{sec:exp_faster}.
% From a memory perspective, \sysname distributes sequences in different workers, and thus evenly distributes the activation memory. Megatron-LM distributes both activations and model weights. For a fair comparison, we analyze~\sysname in tandem with FSDP so that it also distributes model weights. We note that the communication in FSDP is proportional to the size of the model, which is a constant to sequence length. In conclusion, when used with FSDP, ~\sysname achieves up to 4.7x communication volume reduction compared to Megatron-LM with a similar memory footprint, when scaling up the sequence lengths. 
% Note that the difference in communication volume will increase if we further consider models such as GQA or MQA, for which we provide more analysis in~\S~\ref{sec:exp_faster}. 
However, we also note that this is a theoretical analysis, where the wall-clock time may differ because of factors such as implementations. In the experiment section, we provide wall-clock end-to-end results for comparison.

\section{Implementation Details}\label{app:imple_details}
We build the kernel of~\attnname, modifying from the Triton kernel of FlashAttention2 in 500 lines of codes (LoCs). We implement the load balancing and overlapping scheduling n Python and NCCL Pytorch bindings in 1000 LoCs~\citep{paszke2019pytorch, jeaugey2017nccl}, and the checkpointing strategy in 600 lines of Pytorch. %It is attention backend agnostic.% To reduce the memory consumption and reach faster speed in the attention module, we use the FlashAttention2 algorithm~\citep{dao2023flashattention}. We use the triton~\citep{tillet2019triton} implementation and minimally modify it to keep around statistics in the flash attention algorithm. 
We use block sizes of 128 and the number of stages to 1 in the kernel for the best performance in our cluster. %We reuse the C++ backward kernels of FlashAttention2 because we do not need to modify the backward logic. 
We evaluate~\attnname using FSDP (inter-node if applicable) so that it consumes similar memory than the Megatron-LM baseline for a fair comparison~\citep{zhao2023pytorch}.
%to reduce the memory footprint of data parallelism~\citep{zhao2023pytorch}. 
For fair comparisons, we run all comparisons using the same attention backend. We also add support for Megatron-LM so that comparing with them can produce a more insightful analysis: (1) not materializing the causal attention mask, greatly reducing the memory footprint. For instance, without this support, Megatron-LM will run out of memory with LLaMA-7B at a sequence length of 16K per GPU. (2) head padding where the attention heads cannot be divided by device number. All results are gathered with Adam optimizer, 10 iterations of warm-up, and averaged over the additional 10 iterations.

\section{Discussion on sparse attention}
\label{sec:appendix_sparse}
While this paper focuses on discussing the exact attention mechanism, we also provide possible solutions for sparse patterns and hope it can inspire future works. In particular, we discuss load balancing for local sliding windows and global attention~\citep{beltagy2020longformer}.

\textbf{Local sliding windows} For local sliding windows, the workload is naturally (near) balanced, regardless of single directional or bidirectional attention. Thus, simply disregarding the attention logic to non-local workers suffices. For instance, in exact attention, worker 7 needs to compute attention to all other workers. If the sliding window has a number of tokens equal to that of one worker, then worker 7 only needs to attend to itself and tokens in worker 6. In other words, it only needs to fetch key and value from worker 6, and compute attention. In terms of implementation change, the system merely needs to change the end condition of the for loop (from looping worker 1 - worker 7 to looping only from worker 6 - worker 7).

\textbf{Global attention} In global attention, there are a certain number of global tokens that all later tokens need to attend to, which are used to capture the global information. To adapt \sysname to this, one solution is to keep a replica of all the global tokens in each worker, which is simple and practical as otherwise, the global tokens will need to be all-gathered at each time step. The other possibility is to also split the global tokens evenly onto all workers and use all-gather upon computation to further reduce the memory requirement.

\end{document}